\documentclass[sigconf]{acmart}

\usepackage{amsfonts}       
\usepackage{amsmath}
\usepackage{amssymb}
\usepackage{graphicx, subfig}
\usepackage{nicefrac} 
\usepackage[ruled, vlined, linesnumbered]{algorithm2e}
\usepackage{enumerate}
\usepackage{enumitem}
\usepackage{geometry}
\usepackage{tikz}
\usetikzlibrary{shapes.geometric}
\usepackage{arydshln}
\usepackage{xspace}
\usepackage{amssymb}
\usepackage{bbding}
\usepackage{tcolorbox}
\usepackage{multirow}
\usepackage{threeparttable}
\usepackage{soul}
\usepackage{color}

\usepackage[capitalize,noabbrev]{cleveref}
\usepackage{makecell}
\usepackage{wrapfig}
\usepackage{caption}
\usepackage{subcaption}
\usepackage{pifont}
\usepackage{tikz}
\usepackage{hyperref}
\usepackage[thicklines]{cancel}
\SetKwInput{KwInput}{Input} 
\SetKwInput{KwOutput}{Output}   
\usepackage{balance} 

\setulcolor{red}

\def\header{\vspace{2mm} \noindent}

\AtBeginDocument{%
  \providecommand\BibTeX{{%
    \normalfont B\kern-0.5em{\scshape i\kern-0.25em b}\kern-0.8em\TeX}}}

\copyrightyear{2025}
\acmYear{2025}
\setcopyright{acmlicensed}
\acmConference[KDD '25] {Proceedings of the 31st ACM SIGKDD Conference on Knowledge Discovery and Data Mining V.2}{August 3--7, 2025}{Toronto, ON, Canada.}
\acmBooktitle{Proceedings of the 31st ACM SIGKDD Conference on Knowledge Discovery and Data Mining V.2 (KDD '25), August 3--7, 2025, Toronto, ON, Canada}
\acmISBN{979-8-4007-1454-2/25/08}
\acmDOI{10.1145/3711896.3737155}

\begin{document}
\title{TIDFormer: Exploiting Temporal and Interactive Dynamics Makes A Great Dynamic Graph Transformer}

\author{Jie Peng}
\affiliation{
  \institution{Renmin University of China}
  \city{Beijing}\country{China}
  }
\email{peng_jie@ruc.edu.cn}

\author{Zhewei Wei}
\affiliation{%
  \institution{Renmin University of China}
  \city{Beijing}\country{China}
}
\authornote{Zhewei Wei is the corresponding author.}
\email{zhewei@ruc.edu.cn}

\author{Yuhang Ye}
\affiliation{%
  \institution{Huawei}
  \city{Shenzhen, Guangdong}\country{China}
}
\email{yeyuhang@huawei.com}

\begin{abstract}
Due to the proficiency of self-attention mechanisms (SAMs) in capturing dependencies in sequence modeling, several existing dynamic graph neural networks (DGNNs) utilize Transformer architectures with various encoding designs to capture sequential evolutions of dynamic graphs. 
However, the effectiveness and efficiency of these Transformer-based DGNNs vary significantly, highlighting the importance of properly defining the SAM on dynamic graphs and comprehensively encoding temporal and interactive dynamics without extra complex modules.
In this work, we propose \textbf{TIDFormer}, a dynamic graph Trans\textbf{Former} that fully exploits \textbf{T}emporal and \textbf{I}nteractive \textbf{D}ynamics in an efficient manner.
We clarify and verify the interpretability of our proposed SAM, addressing the open problem of its uninterpretable definitions on dynamic graphs in previous works. 
To model the temporal and interactive dynamics, respectively, we utilize the calendar-based time partitioning information and extract informative interaction embeddings for both bipartite and non-bipartite graphs using merely the sampled first-order neighbors.
In addition, we jointly model temporal and interactive features by capturing potential changes in historical interaction patterns through a simple decomposition.
We conduct extensive experiments on several dynamic graph datasets to verify the effectiveness and efficiency of TIDFormer. 
The experimental results demonstrate that TIDFormer excels, outperforming state-of-the-art models across most datasets and experimental settings.
Furthermore, TIDFormer exhibits significant efficiency advantages compared to previous Transformer-based methods.

\end{abstract}

\ccsdesc[500]{Computing methodologies~Artificial intelligence}
\ccsdesc[500]{Information systems~Data mining}

\vspace{-10pt}

\keywords{Dynamic Graph Representation Learning; Transformer; Temporal and Interactive Dynamics}
\vspace{-10pt}

\maketitle

\section{Introduction}\label{intro}
Graph-structured data is ubiquitous, encompassing a wide range of domains such as social networks \cite{DBLP:conf/kdd/KumarZL19,DBLP:conf/wsdm/Song0WCZT19,alvarez2021evolutionary}, recommendation systems \cite{zhang2022dynamic,gao2022graph,TGB-seq}, and traffic networks \cite{xie2020sast,bui2022spatial,sharma2023graph}. 
In the past few years, research has primarily focused on static graphs \cite{kipf2016semi,chen2020simple,peng2024beyond}, which represent fixed node \& edge features and graph structures. 
However, many real-world graph datasets are inherently dynamic, evolving over time with changes in nodes, edges, and their attributes \cite{DBLP:journals/jmlr/KazemiGJKSFP20}. 
These dynamic characteristics make dynamic graphs a crucial field for deeper investigation.

The existing methods of modeling dynamic graphs can be broadly categorized into two types \cite{zheng2024survey}: discrete-time dynamic graph (DTDG) learning \cite{DBLP:conf/wsdm/SankarWGZY20,DBLP:journals/corr/abs-2111-10447,DBLP:conf/kdd/YouDL22} and continuous-time dynamic graph (CTDG) learning \cite{DBLP:conf/iclr/WangCLL021,DBLP:conf/nips/SouzaMKG22,cong2023do}. 
DTDGs evolve in distinct, separate time steps, represented as a series of static graph snapshots, each corresponding to a specific timestamp. 
In contrast, CTDGs represent the evolution of the graph as a continuous process without discrete time steps. 
The addition and deletion of edges or the changes of node attributes are represented as a series of graph events in chronological order, allowing for finer representations of dynamic evolution. 
In this work, we focus on CTDGs due to their ability to offer detailed temporal and interactive dynamics of the graphs. 
In recent years, CTDG learning has received extensive attention. 
Although many CTDG methods have been proposed, including temporal point process-based \cite{trivedi2017know,han2019graph,zhao2023time} methods, memory-based methods \cite{DBLP:journals/corr/abs-2006-10637,wang2021apan,su2024pres}, and random walk-based \cite{DBLP:conf/iclr/WangCLL021,li2023zebra,lu2024improving} methods, they either suffer from high computational costs caused by complex models or poor performance due to simplistic designs.
Consequently, current CTDG learning methods remain unsatisfactory.

Recently, due to the proficiency of self-attention mechanisms (SAMs) in learning dependencies in sequence modeling, several existing DGNNs \cite{DBLP:journals/corr/abs-2006-10637,DBLP:conf/iclr/XuRKKA20,yu2023towards} utilize Transformer architectures with various encoding designs to capture sequential evolutions of dynamic graphs.
However, the definitions of these SAMs vary widely without reasonable interpretability.
For instance, TGAT \cite{DBLP:conf/iclr/XuRKKA20} initially defines a SAM on dynamic graphs at single-node level, that is, learning the dependencies between the entities in the sequence corresponding to a single given node.
The SAM at single-node level fails to interpretably capture the node-pair information required for link-wise downstream tasks.
Subsequent Transformer-based models, such as DyGFormer \cite{yu2023towards}, adjust the SAM without clear explanations. 
Specifically, DyGFormer mixes the sequences of source and target nodes, calculating the attention weights between the entities in the mixed sequence, thereby defining the SAM at multi-node level. 
The SAM at multi-node level could compromise the learning process of the attention weights, as the relationships and chronological orders within and across the sequences of the source and target nodes differ significantly.
Hence, there is a need for an \textbf{interpretable SAM} that is correctly defined according to the requirements of link-wise downstream tasks and the correct chronological characteristics of dynamic graphs.

In addition to the definition of SAMs, several modules have been designed to encode temporal and interactive dynamics, respectively.
For encoding temporal dynamics, \emph{the temporal information of each edge}, many methods use the cosine function to encode the original fine-grained timestamps of dynamic graphs, distinguishing the nodes of recent times from those of distant times. 
However, this cosine-based mapping can not capture the phased and periodic characteristics that are common in time-series data like dynamic graphs.
Although FreeDyG \cite{tian2023freedyg} attempts to learn potential patterns in the frequency domain, it introduces additional computational costs, such as the calculation of the Fourier Transform \cite{sneddon1995fourier}. 
Thus, effectively depicting \textbf{temporal dynamics} without adding complex modules in the encoding stage remains an open question.

In the process of encoding interactive dynamics, \emph{the interaction evolution between nodes over time}, most existing CTDG methods only consider the representations within the sequence of individual nodes, neglecting the interactive information between node pairs. 
For example, they often overlook the implicit relationship between the interacted node pair in link prediction tasks. 
To address this issue, both DyGFormer \cite{yu2023towards} and FreeDyG \cite{tian2023freedyg} propose similar neighbor interaction encoding modules to exploit the correlation between node pairs. 
However, these neighbor interaction encoding modules have significant drawbacks when dealing with dynamic bipartite graphs, such as user-item dynamic graphs in recommendation systems, since the source node and the target node have no common neighbors under first-order sampling.
As a result, the module degenerates into merely recording the historical interaction times of node pairs.
Therefore, the \textbf{interactive dynamics} encoding process warrants further improvement.

To sum up, the current Transformer-based CTDG models face the following two key problems:

\textbf{1.} \emph{Uninterpretable definitions of the SAM on dynamic graphs.}

\textbf{2.} \emph{Insufficient and inefficient modeling of temporal and interactive dynamics.}

To address these problems, we propose \textbf{TIDFormer}, a dynamic graph Trans\textbf{Former} that fully exploits \textbf{T}emporal and \textbf{I}nteractive \textbf{D}ynamics with an interpretable SAM for modeling dynamic graphs.

Considering the link-wise downstream tasks and the chronological characteristics of dynamic graphs, we argue that the SAM on dynamic graphs should be built at {interaction level}.
The SAM at {interaction level} interpretably captures the dependencies between node pairs in the sequences of historical interactions without compromising the original chronological order.
Each entity in the reconstructed sequence consists of both its own embeddings and the interactive information of the interacted source or target nodes.
This SAM effectively models the temporal and interactive dynamics, incorporating interactive information between node pairs. 
With regards to the interpretability evaluation of different types of SAMs, we conduct experimental analysis to demonstrate the strong correlation between the high-frequency interaction nodes and the corresponding attention weights learned by the SAM at interaction level, which verifies the interpretability of our proposed SAM.
Thus, we achieve an \textbf{interpretable SAM at interaction level}, addressing the open problem of its uninterpretable definitions on dynamic graphs in previous works.

To fully exploit the temporal and interactive dynamics, we propose a series of simple yet effective encoding modules.
For modeling temporal dynamics, we propose a \textbf{Mixed-granularity Temporal Encoding} module that considers both the fine-grained timestamp information and the coarse-grained time partitioning information of the calendar. 
These weekly, monthly, and yearly timestamps are highly informative for downstream tasks but have been largely overlooked in previous DGNNs.
For modeling interactive dynamics, we propose a \textbf{Bidirectional Interaction Encoding} module, which addresses the problem of invalid neighbor encoding for bipartite graphs in previous works.
It uses a bidirectional mechanism that efficiently reconstructs the sequence with merely first-order sampling.
This module significantly enhances the utilization of interactive information by increasing the receptive fields of source and target nodes without additional sampling of higher-order neighbors.
Furthermore, we propose a \textbf{Seasonality \& Trend Encoding} module, which effectively captures potential changes in historical interaction patterns over time via a simple decomposition. 
Specifically, this module decomposes the input sequence using an average pooling operation, thereby extracting trend and seasonal features that jointly model both temporal and interactive dynamics.

We conduct extensive experiments on seven real-world CTDG datasets to evaluate the effectiveness and efficiency of TIDFormer, including experiments on link prediction and node classification tasks under various settings.
The experimental results demonstrate that TIDFormer excels, outperforming state-of-the-art (SOTA) models across most datasets and experimental settings.
Furthermore, TIDFormer exhibits significant efficiency advantages compared to previous Transformer-based methods.
This clearly shows that exploiting temporal and interactive dynamics can easily make a great dynamic graph Transformer.

Our main contributions are summarized as follows:
\begin{itemize}
    \item \textbf{TIDFormer.} 
    We propose TIDFormer, a dynamic graph Transformer that fully mines temporal and interactive dynamics with an interpretable SAM.
    \item \textbf{Interpretable SAM.} 
    We propose an interpretable SAM at interaction level, addressing the open problem of previous uninterpretable definitions.
    \item \textbf{Exploiting temporal and interactive dynamics.} 
    We novelly introduce a series of simple yet effective encoding modules to capture the temporal and interactive dynamics. 
\end{itemize}

\section{Related Work}
{\textbf{Dynamic Graph Learning.}}
The existing methods of modeling dynamic graphs can be broadly categorized into two types: discrete-time dynamic graph (DTDG) learning \cite{DBLP:conf/wsdm/SankarWGZY20,DBLP:journals/corr/abs-2111-10447,DBLP:conf/kdd/YouDL22} and continuous-time dynamic graph (CTDG) learning \cite{DBLP:conf/iclr/WangCLL021,DBLP:conf/nips/SouzaMKG22,cong2023do}. 
DTDGs evolve in distinct, separate time steps, represented as a series of static graph snapshots, each corresponding to a specific timestamp. 
For instance, WD-GCN \cite{manessi2020dynamic} is one of the earliest algorithms that combine Graph Neural Networks and Recurrent Neural Networks for DTDG learning, integrating GCN \cite{kipf2016semi} and LSTM \cite{hochreiter1997long} models.
In contrast, CTDGs represent the evolution of the graph as a continuous process without discrete time steps. 
The addition and deletion of edges or the changes in node attributes are represented as a series of graph events in chronological order, allowing for a finer and more accurate representation of dynamic evolution. 
Several CTDG models are designed with temporal point process-based methods \cite{trivedi2017know,han2019graph,zhao2023time}, memory-based methods \cite{DBLP:journals/corr/abs-2006-10637,wang2021apan,su2024pres}, and random walk-based methods \cite{DBLP:conf/iclr/WangCLL021,li2023zebra,lu2024improving}.
Recently, another line of work has focused on encoding temporal or interactive information.
For instance, TGAT \cite{DBLP:conf/iclr/XuRKKA20} uses the SAM and a time encoder to create the TGAT layer, achieving similar propagation capabilities as Graph Attention Networks \cite{velivckovic2017graph} by stacking TGAT layers. 
DyGFormer \cite{yu2023towards} proposes a Transformer-based dynamic graph learning architecture that can capture neighbor information of node pairs. 
FreeDyG \cite{tian2023freedyg} introduces an interaction frequency encoding module via Fourier Transform that explicitly captures node pair interaction frequencies to address the shift phenomenon.
TPNet \cite{lu2024improving} implicitly computes temporal walk matrices that incorporate the time decay effect to model temporal and structural information.
Although the above models promote the development of CTDG learning, they either suffer from high computational costs caused by complex models or poor performance due to insufficient modeling of temporal and interactive dynamics, especially for those Transformer-based methods.
Consequently, current CTDG methods remain unsatisfactory.

\header{\textbf{Graph Transformer.}}
Transformer has revolutionized natural language processing by leveraging SAMs to model dependencies across sequences. 
Its ability to handle long-range dependencies and enable parallel training has made it a dominant architecture in various sequence-based tasks.
Hence, in graph learning, several models have emerged that integrate Transformers with static graph data to harness the strengths of both. 
Notable examples include Graphormer \cite{ying2021transformers}, which introduces positional and structural encodings specifically designed for graphs, allowing the Transformer to process graph data effectively. 
PolyFormer \cite{ma2023polyformer} proposes a tailored polynomial SAM that serves as a node-wise graph filter, offering powerful representation capabilities.
Recently, researchers have focused on merging Transformers with dynamic graphs since their sequential structure naturally aligns with SAMs to capture long-range dependencies.
For instance, TGN \cite{DBLP:journals/corr/abs-2006-10637}, TGAT \cite{DBLP:conf/iclr/XuRKKA20}, and DyGFormer \cite{yu2023towards} all incorporate Transformer's SAM to handle the evolution of dynamic graphs. 
However, the definitions of their SAMs vary widely without reasonable interpretability, hindering the development of Transformer-based dynamic graph learning.
More recently, SimpleDyG \cite{wu2024feasibility} uses several complex tokenization methods from natural language processing to generate new sequences of temporal ego-graphs, which complicates the overall process.
Hence, it is necessary to define an interpretable SAM according to link-wise downstream tasks and the chronological characteristics of dynamic graphs.
Meanwhile, it is still an open question of how to effectively depict temporal and interactive dynamics without complex modules in the encoding stage.

\section{Preliminaries}
\textbf{Notations.}
We consider a dynamic graph $G=(V,E,T)$ with sets of nodes $V$, edges $E$, and timestamps $T$.
A CTDG $G$ describes a continuous graph evolving process that can be formulated as a series of sequential graph events $\{Event_1, \ldots, Event_N\}$, where $N$ is the number of graph events.
Specifically, each graph event denotes an interaction between a source node $src_i\in V$ and a target node $tgt_i\in V$ at time $t_i\in T$ as $\left(src_i,tgt_i,t_i\right)$, where $1\leq i \leq N$.
Nodes $src_i$ and $tgt_i$ have node feature $\mathcal{N}_{i}^{src},\mathcal{N}_{i}^{tgt}\in\mathbb{R}^{d_n}$, and the interaction between them at $t_i$ has an edge feature $\mathcal{E}_{i}^{t_i}\in\mathbb{R}^{d_e}$, where $d_n$ and $d_e$ denote the dimensions of the node and edge embeddings. 
Hence, the dynamic graph $G$ could be further represented as a sequence of chronologically ordered historical interactions $G=\{\left(src_1,tgt_1,t_1\right), \ldots, \left(src_N,tgt_N,t_N \right)\}$ with $0\leq t_1\leq \ldots \leq t_N$.

\header{\textbf{Downstream Tasks.}}
Consistent with earlier works \cite{DBLP:conf/iclr/XuRKKA20,DBLP:journals/corr/abs-2006-10637,yu2023towards}, we select two crucial tasks in dynamic graph learning to verify the effectiveness of TIDFormer: 
(1) \emph{Dynamic Link Prediction}: Predicting whether the source node and the target node are connected at a given time;
(2) \emph{Dynamic Node Classification}: Determining the state of the node at a given time.

\begin{figure*}[t]
\vskip -0.1in
    \centering
    \subfloat[Single-node Level]{\includegraphics[height=18mm]{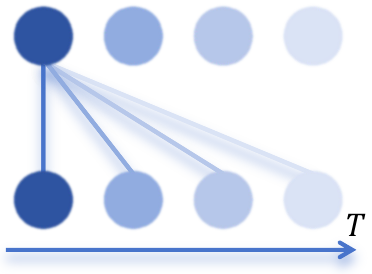}}\hspace{13mm}
    \subfloat[Multi-node Level]{\includegraphics[height=18mm]{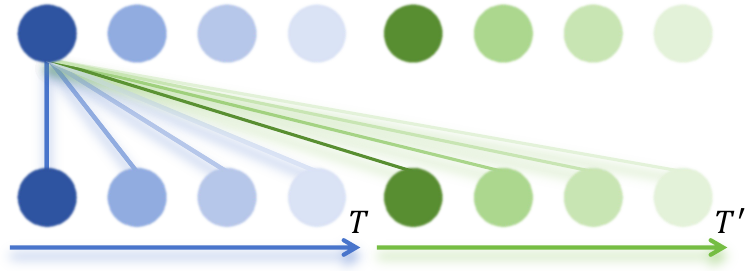}}\hspace{13mm}
    \subfloat[Interaction Level]{\includegraphics[height=18mm]{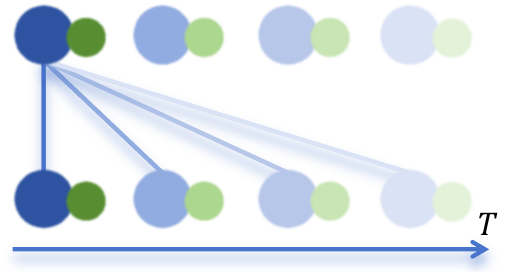}}\\
    \vskip -0.1in
    \caption{Illustration of three types of SAMs on dynamic graphs at different levels.}
    \label{fig:SAM}
    \vskip -0.2in
\end{figure*}

\begin{figure*}[t]
    \centering
    \subfloat[SAM at SL (TGAT)]{\includegraphics[height=27mm]{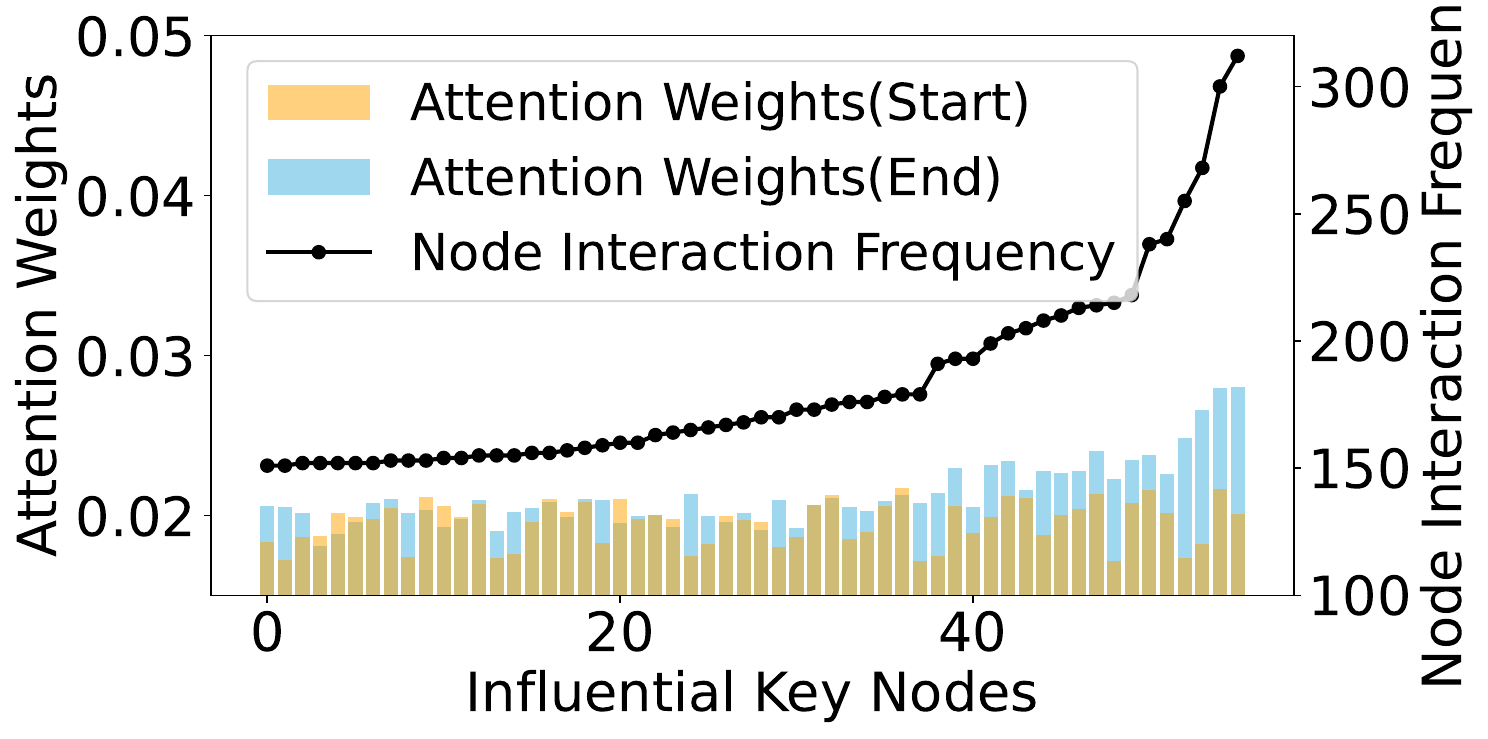}}\hspace{6mm}
    \subfloat[SAM at ML (DyGFormer)]{\includegraphics[height=27mm]{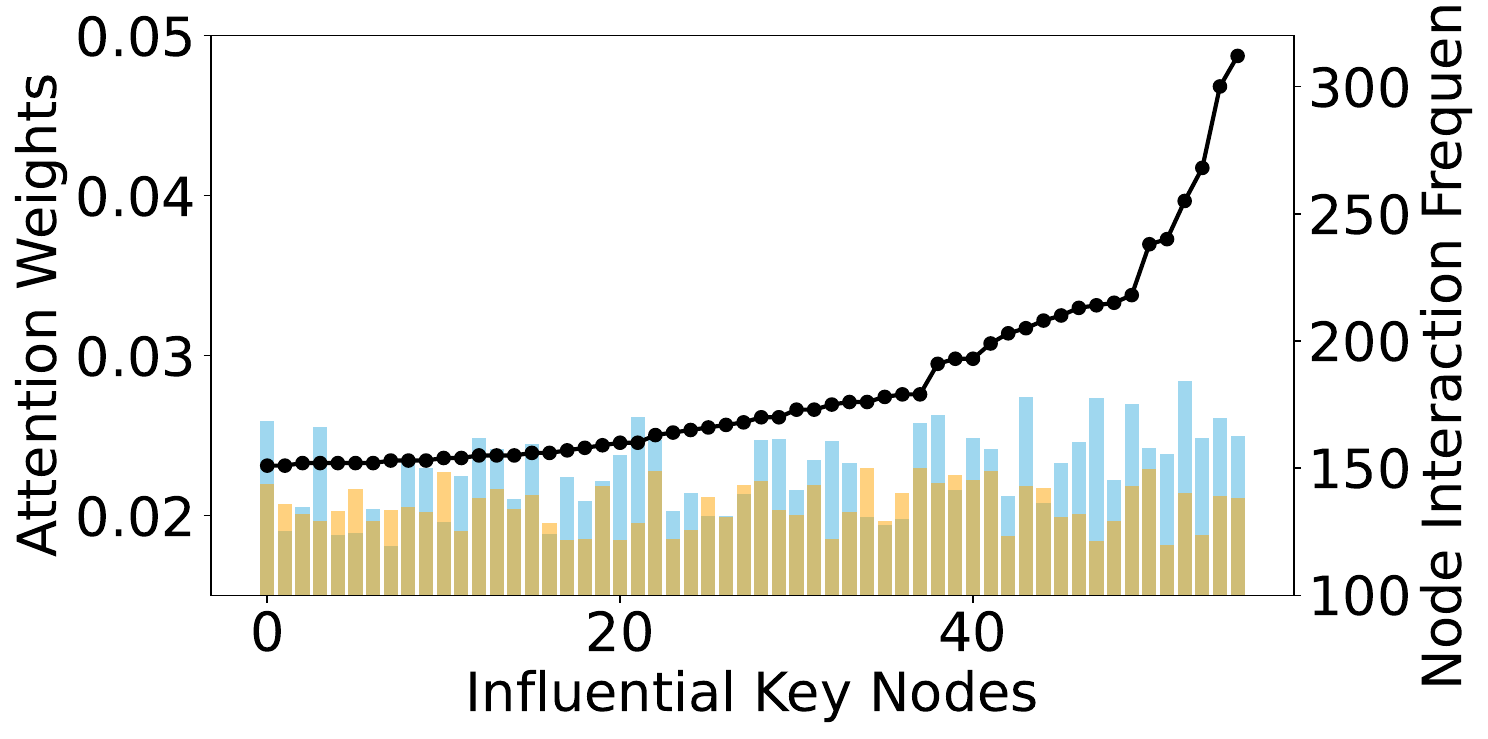}}\hspace{6mm}
    \subfloat[SAM at IL (TIDFormer)]{\includegraphics[height=27mm]{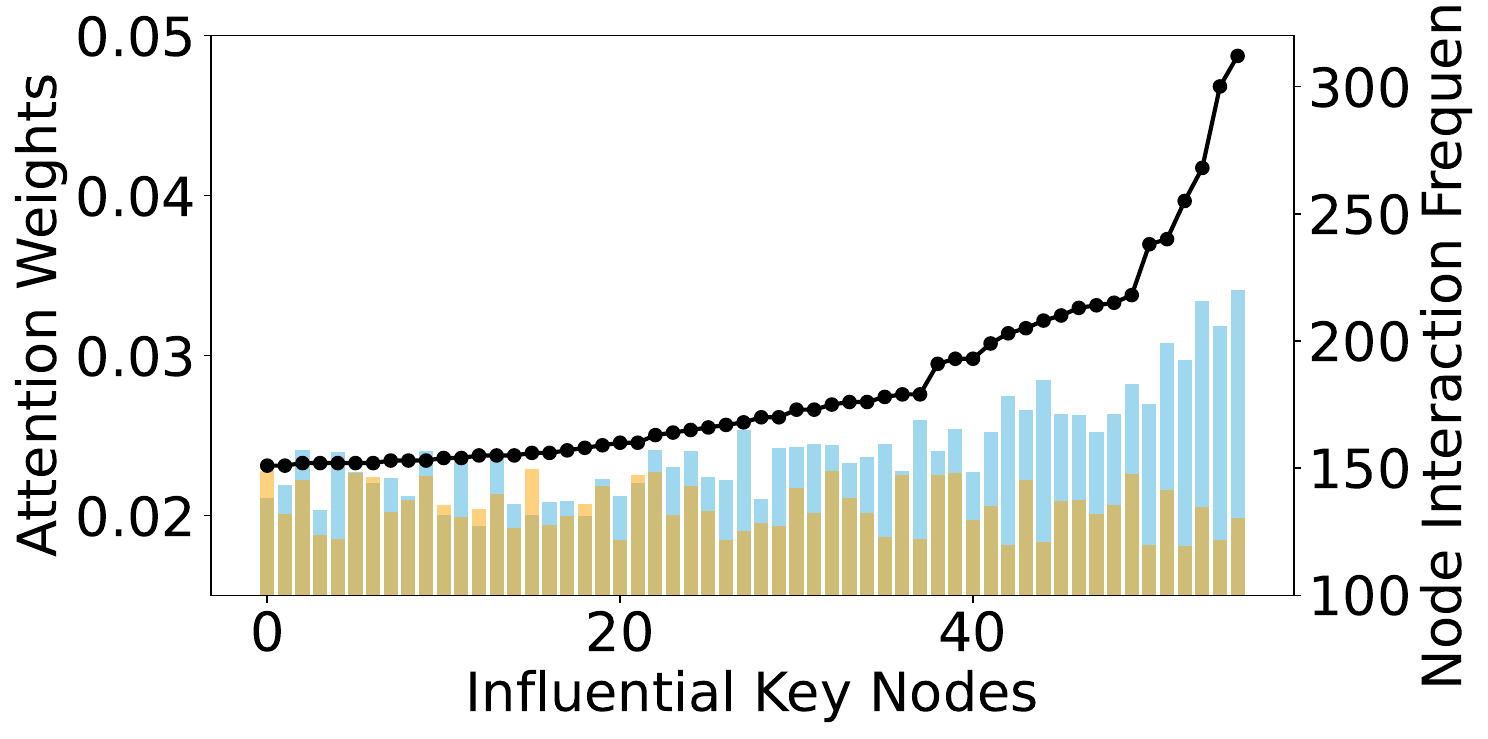}}\\
    \vskip -0.1in
    \caption{Illustration of the key nodes' attention weights of three types of SAMs from the start to the end of training on MOOC.}
    \label{fig:SAM attention weights}
    \vskip -0.1in
\end{figure*}

\section{Metholodgy}
\subsection{Interpretable SAM on Dynamic Graphs}\label{three SAMs}
Since Transformer's SAM excels at learning dependencies in sequence modeling, several existing DGNNs use Transformer as the backbone model, employing various encoding designs to tokenize the sequential evolutions of dynamic graphs.
The SAM in the vanilla Transformer can be decomposed into two stages:
(1) \emph{Tokenization of the sequence}, generating an ordered sequence $X\in\mathbb{R}^{n\times (d+d_{p})}$ with a fixed or learnable position embedding $\mathcal{P}\in\mathbb{R}^{d_{p}}$ as shown in \cref{equ:vanilla tokenization}, where $n$, $d$ and $d_p$ denote the length of the sequence, the dimensions of the original sequence and position embeddings; 
(2) \emph{Calculation of the attention weights}, obtaining the scaled dot-product attention weights with learnable projection matrices $W_Q\in\mathbb{R}^{(d+d_{p})\times d_k}$, $W_K\in\mathbb{R}^{(d+d_{p})\times d_k}$, and $W_V\in\mathbb{R}^{(d+d_{p})\times d_v}$ as shown in \cref{equ:vanilla attention}.
\begin{equation}\label{equ:vanilla tokenization}
    X=[\mathcal{X}_1||\mathcal{P}_1, \mathcal{X}_2||\mathcal{P}_2, \ldots,\mathcal{X}_n||\mathcal{P}_n]\in\mathbb{R}^{n\times (d+d_{p})},
\end{equation}
where $\mathcal{X}=[\mathcal{X}_1, \mathcal{X}_2, \ldots, \mathcal{X}_n]\in\mathbb{R}^{n\times d}$ is the original sequence.
\begin{equation}\label{equ:vanilla attention}
    Attention(Q,K,V)=Softmax\left(\frac{(XW_Q)(XW_K)^{\top}}{\sqrt{d_{k}}}\right)XW_V.
\end{equation}

In previous works, when SAMs meet the sequential data of dynamic graphs, different tokenization strategies are proposed at single-node level, multi-node level, and so on.

\header{\textbf{Single-node Level (SL).}}
The SAM in TGAT \cite{DBLP:conf/iclr/XuRKKA20} is defined at SL. 
TGAT replaces the original position embedding $\mathcal{P}$ with a temporal embedding $\mathcal{T}\in\mathbb{R}^{d_{t}}$ to obtain a chronologically ordered sequence ${H}_{SL}\in\mathbb{R}^{n\times (d+d_{t})}$ as shown in \cref{equ:single-node tokenization}, where $d_t$ denotes the dimension of temporal embeddings.
The SAM at SL is illustrated in \cref{fig:SAM}(a).
\begin{equation}\label{equ:single-node tokenization}
    H_{SL}=[\mathcal{H}_1||\mathcal{T}_1, \mathcal{H}_2||\mathcal{T}_2, \ldots,\mathcal{H}_n||\mathcal{T}_n]\in\mathbb{R}^{n\times (d+d_{t})},
\end{equation}
where $\mathcal{H}=[\mathcal{H}_1, \mathcal{H}_2, \ldots, \mathcal{H}_n]\in\mathbb{R}^{n\times d}$ consists of node features $\mathcal{N}\in\mathbb{R}^{n\times d_n}$ and edge features $\mathcal{E}\in\mathbb{R}^{n\times d_e}$ of a single given source node or target node, with $d=d_n+d_e$.
The SAM at SL models the representation of the node independently.
Thus, it cannot capture the interactive information between the sequences of source and target nodes when processing downstream tasks such as link prediction. 
As a result, there is a lack of interpretability between the SAM at SL and link-wise downstream tasks.

\header{\textbf{Multi-node Level (ML).}}
The SAM in DyGFormer \cite{yu2023towards} is defined at ML.  
Based on the SAM at SL, DyGFormer constructs a mixed sequence $H_{ML}\in\mathbb{R}^{2n\times(d+d_{t})}$ as shown in \cref{equ:multi-node tokenization}, which combines two chronologically ordered sequences of source and target nodes, respectively, aiming to learn the temporal dependencies within and across two sequences.
The illustration of the mechanism at ML is shown in \cref{fig:SAM}(b).
\begin{equation}\label{equ:multi-node tokenization}
\begin{aligned}
    H_{ML}=[&\mathcal{H}_1^{src}||\mathcal{T}_1^{src}, \mathcal{H}_2^{src}||\mathcal{T}_2^{src}, \ldots,\mathcal{H}_n^{src}||\mathcal{T}_n^{src},\\
    &\mathcal{H}_1^{tgt}||\mathcal{T}_1^{tgt}, \mathcal{H}_2^{tgt}||\mathcal{T}_2^{tgt}, \ldots,\mathcal{H}_n^{tgt}||\mathcal{T}_n^{tgt}]\in\mathbb{R}^{2n\times (d+d_{t})},
\end{aligned}
\end{equation}
where $\mathcal{H}_i^{src}$ and $\mathcal{H}_i^{tgt}$ denote the $i$-th token in the sequences of source and target nodes, respectively, and $1\leq i \leq n$.
However, the mixed sequence after tokenization in the SAM at ML could compromise the correct chronological order to some extent, particularly when calculating the attention weights between tokens belonging to the source node and the target node, as their chronological orders are different and independent ($T$ for source nodes and $T^{\prime}$ for target nodes in \cref{fig:SAM}(b)). 
As a result, there is a lack of interpretability between the SAM at ML and the natural correct chronological order of the dynamic graph. 
Therefore, it is necessary to further refine the SAM for dynamic graphs in terms of interpretability.

\header{\textbf{Interaction Level (IL).}}
Based on the shortcomings of the above two types of SAMs, we argue that the SAM on dynamic graphs should be defined in an interpretable manner while considering the link-wise downstream tasks and the chronological characteristics.
Thus, we propose the SAM at IL, which interpretably captures the dependencies between node pairs in the sequences of historical interactions without compromising the original chronological order.
Specifically, the SAM at IL is based on a well-designed sequence $H_{IL}\in\mathbb{R}^{n\times(d+d_{i})}$ as shown in \cref{equ:interaction tokenization}.
$H_{IL}$ integrates the original node and edge features, with a new temporal-interactive embedding $\mathcal{I}\in\mathbb{R}^{n\times d_{i}}$ between the source and target nodes in the original chronological order, where $d_i$ denotes the dimension of temporal-interactive embedding.
The structure of the temporal-interactive embedding $\mathcal{I}$ will be further explained in \cref{sec:TIDFormer}.
The illustration of the SAM at IL is shown in \cref{fig:SAM}(c).
\begin{equation}\label{equ:interaction tokenization}
    H_{IL}=[\mathcal{H}_1||\mathcal{I}_1, \mathcal{H}_2||\mathcal{I}_2, \ldots,\mathcal{H}_n||\mathcal{I}_n]\in\mathbb{R}^{n\times (d+d_{i})}.
\end{equation}

Therefore, the proposed SAM at IL addresses the lack of interpretability between SAMs and link-wise downstream tasks or chronological characteristics, two issues revealed by previous dynamic graph models' SAMs during tokenization.
In \cref{sec:inter analysis}, we will further evaluate the interpretability of the SAM at IL by analyzing the calculated attention weights during training.

\subsection{Interpretability Analysis of SAMs}\label{sec:inter analysis}
Previous Transformer-based dynamic graph methods primarily derive node representations by modeling the relationships between nodes within the sampled neighbor sequence through the SAM's attention weights. 
Naturally, the SAM's learnable attention weights intuitively reflect the significance of each node to the current modeling process. 
Specifically, nodes with high interaction frequency could exert greater influence and tend to receive higher attention weights. 
Therefore, a well-defined and properly designed SAM for dynamic graphs is expected to effectively identify influential key nodes and interpretably learn higher attention weights for these nodes.
To assess the interpretability of SAMs, we select nodes with interaction frequency greater than 150 as \emph{influential key nodes} and track their corresponding attention weights during training on the MOOC dataset \cite{poursafaei2022towards}. 
Nodes that exhibit active interactions over time are expected to show increasing influence (i.e., higher attention weights) throughout the modeling process, as they hold greater importance in predicting future links.

The results presented in \cref{fig:SAM attention weights} reveal that the initial attention weights follow similar random distributions at the start of training (the orange parts). 
However, by the end of the training, the distributions of the learned attention weights among the three types of SAMs differ significantly (the blue parts).
For the SAM at ML, the final attention weights show no significant correlation with the node interaction frequency, which could be due to the chronologically disordered sequence in DyGFormer. 
In contrast, for the SAMs at SL and IL, the attention weights are correlated with node interaction frequencies: influential key nodes with higher interaction frequency learn higher attention weights.
However, since our proposed SAM at IL incorporates interaction information between node pairs during the tokenization process, it has a stronger ability to identify influential key nodes, resulting in better interpretability in learning attention weights for nodes with active interactions.

In summary, previous methods struggle to accurately capture high-frequency interaction patterns, resulting in either less relevant relationships (SAM at SL) between the learned attention weights and node interaction frequencies or greater fluctuation of the learning outcomes (SAM at ML).
In contrast, the SAM at IL demonstrates superior interpretability in capturing influential key nodes and dynamic behaviors.

\begin{figure*}[t]
    \centering
    \includegraphics[width=2.0\columnwidth]{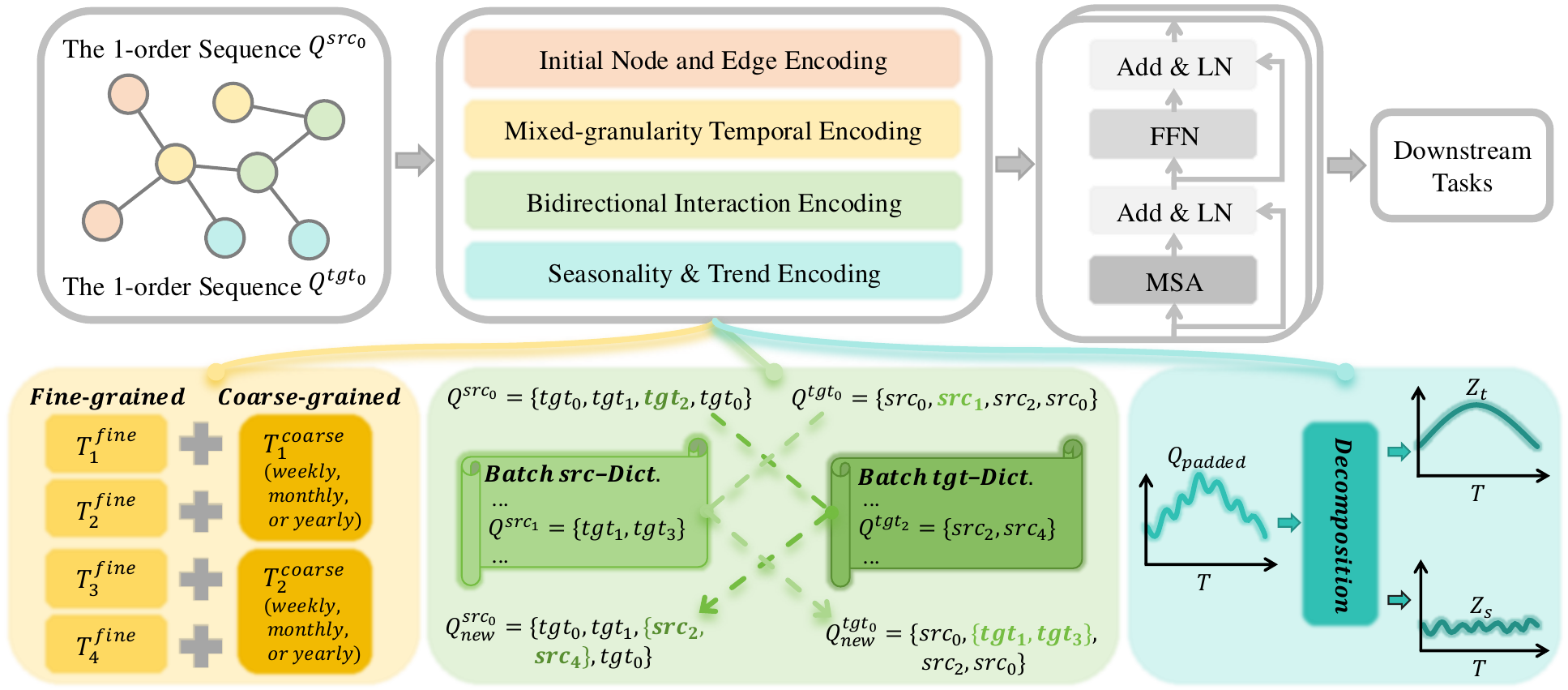}
    \caption{The pipeline of TIDFormer.}
    \label{fig:pipeline}
\end{figure*}

\subsection{TIDFormer}\label{sec:TIDFormer}
In this work, we propose TIDFormer, a Transformer-based dynamic graph model that fully exploits temporal and interactive dynamics with an interpretable SAM at IL.
Consistent with the basic settings of previous works \cite{yu2023towards,tian2023freedyg}, we sample $n$ first-order neighbors of source and target nodes in chronological order to obtain the initial sequences, respectively, which contain the neighbor node embeddings and edge embeddings.
For the encoding stage of TIDFormer, we then design a \emph{Mixed-granularity Temporal Encoding} module to merge the original fine-grained timestamp information with the coarse-grained time partitioning information of the calendar.
Additionally,  we propose a \emph{Bidirectional Interaction Encoding} module to exploit more informative interaction embeddings for both bipartite and non-bipartite graphs using merely first-order neighbors.
Furthermore, we introduce a \emph{Seasonality \& Trend Encoding} module to capture potential changes of historical interaction patterns over time via a simple decomposition.
After the above tokenization and encoding, the well-designed sequence is fed into the Transformer with the interpretable SAM at IL for capturing temporal and interactive dependencies. 
Finally, the representations of the source node and the target node at a given time can be obtained, which are used for downstream tasks.
The pipeline of TIDFormer is shown in \cref{fig:pipeline}.

\header{\textbf{Initial Node and Edge Encoding.}}
Based on the sampled $n$ first-order neighbor nodes, we directly retrieve the neighbor node embeddings $\mathcal{N}\in\mathbb{R}^{n\times d_{n}}$ and the corresponding edge embeddings $\mathcal{E}\in\mathbb{R}^{n\times d_{e}}$ to construct the initial sequence $\mathcal{H}=[\mathcal{N},\mathcal{E}]\in\mathbb{R}^{n\times(d_{n}+d_{e})}$, with zero-padding guaranteeing the length of the sequence.

\header{\textbf{Mixed-granularity Temporal Encoding (MTE).}}
When modeling real-world sequential data with temporal features, time partitioning information, such as weekly, monthly, and yearly calendar time segments, is usually accessible. 
These time segments are highly informative but are rarely utilized in modeling the temporal features of dynamic graphs \cite{DBLP:conf/iclr/XuRKKA20,cong2023do}. 
Hence, we propose the MTE module, which integrates the fine-grained timestamp information with the coarse-grained time partitioning information of the calendar. 
The illustration of the MTE module is shown in \cref{fig:pipeline}.

The original fine-grained relative timestamp interval $\Delta \mathcal{T}_i^{fine}$ is calculated as $\Delta \mathcal{T}_i^{fine}=\mathcal{T}_0^{fine}-\mathcal{T}_i^{fine}$, where $\mathcal{T}_0^{fine}$ denotes the timestamp at which we conduct the downstream task and $\mathcal{T}_i^{fine}$ denotes the $i$-th timestamp of the sequential data, with $1\leq i\leq n$. 
Following \cite{cong2023do}, $\Delta \mathcal{T}_i^{fine}$ is encoded by $\cos(\omega\Delta \mathcal{T}_i^{fine})$, where $\omega=\{\alpha^{-(j-1)/\beta}\}^{d_t}_{j=1}$ is a weighted constant vector to better optimize the training process \cite{cong2023do}.
Hyperparameters $\alpha$, $\beta$ are selected to guarantee $\Delta \mathcal{T}_{max}^{fine}\times\alpha^{-(j-1)/\beta}\rightarrow 0$ as $i\rightarrow d_{t}$ to distinguish all timestamps.

For the coarse-grained time information, we segment the original time using weekly, monthly, or yearly real-world calendar information, depending on the time span of the dataset. 
For example, while the Wikipedia dataset \cite{poursafaei2022towards} spans a month, we can use the weekly calendar information to divide the raw data into $4$ weekly segments to capture a more global time dependency. 
Thus, the time segment $R$ of the Wikipedia dataset is $4$.
Mathematically, the calculation of the coarse-grained relative timestamp interval $\Delta \mathcal{T}_i^{coarse}$ is shown in \cref{equ:coarse-grained relative timestamp}.
\begin{equation}\label{equ:coarse-grained relative timestamp}
\begin{aligned}
   &\Delta \mathcal{T}_i^{coarse-weekly}=\Delta \mathcal{T}_i^{fine}//(24\times7\times3,600~\text{sec}),\\
   &\Delta \mathcal{T}_i^{coarse-monthly}=\Delta \mathcal{T}_i^{fine}//(24\times7\times30\times3,600~\text{sec}),\\
   &\Delta \mathcal{T}_i^{coarse-yearly}=\Delta \mathcal{T}_i^{fine}//(24\times7\times365\times3,600~\text{sec}).
\end{aligned}
\end{equation}

We then encode $\Delta \mathcal{T}_i^{coarse}$ by $\theta\Delta \mathcal{T}_i^{coarse}$, where $\theta=\{1/R\}^{d_t}$.
Finally, based on the multi-grained temporal information, we obtain the MTE embeddings $\mathcal{T}^{mix}=[\mathcal{T}_1^{mix}, \ldots, \mathcal{T}_i^{mix}, \ldots, \mathcal{T}_n^{mix}]\in\mathbb{R}^{n\times d_{t}}$ by
\begin{equation}\label{equ:mixed-granularity temporal embedding}
\begin{aligned}
    &\mathcal{T}_i^{mix}=\cos(\omega\Delta \mathcal{T}_i^{fine})+\theta\Delta \mathcal{T}_i^{coarse}\in\mathbb{R}^{d_{t}},~or\\
    &\mathcal{T}_i^{mix}=\cos(\omega\Delta \mathcal{T}_i^{fine})~\|~\theta\Delta \mathcal{T}_i^{coarse}\in\mathbb{R}^{d_{t}\times 2}.
\end{aligned}
\end{equation}

The MTE module works as enhanced position embeddings, providing local time information via fine-grained temporal embeddings (\emph{distinctive time embedding for each timestamp}) and global time information through coarse-grained temporal embeddings (\emph{identical time embedding for each timestamp in the same time segment}).

\header{\textbf{Bidirectional Interaction Encoding (BIE).}}
Recently, researchers have paid more attention to modeling the interactive information between source and target nodes, proposing various modules that encode common neighbors or interaction frequencies \cite{yu2023towards,tian2023freedyg,zhang2024efficient}. 
However, these methods either face degradation problems when dealing with bipartite dynamic graphs since the source node and the target node have no common neighbors under first-order sampling or incur high computational costs when sampling higher-order neighbors.
Naturally, our design intuition is to address the degradation issues while avoiding the high costs of simply sampling higher-order neighbors.
Hence, we propose the BIE module, which uses a bidirectional mechanism that efficiently reconstructs the sequence with merely first-order sampling.
The illustration of the process is shown in \cref{fig:pipeline}.

Consider two sequences $Q^{src_0}$ and $Q^{tgt_0}$, which consist of the sampled $n$ first-order neighbor nodes of the source node $src_0$ and the target node $tgt_0$, respectively.
During training, we adopt the batch processing method and set the batch size to $m$, maintaining two dynamic first-order neighbor dictionaries, \emph{src-Dict} and \emph{tgt-Dict}, which store the $m$ source node sequences and the $m$ target node sequences, respectively.
We then perform a bidirectional conversion on $Q^{src_0}$ and $Q^{tgt_0}$ to reconstruct them.
For simplicity, we denote $Q^{src_0}\cup Q^{tgt_0}$ as $\Psi$ and $Q^{src_0}\cap Q^{tgt_0}$ as $\Phi$.
Formally, for each node of $Q^{src_0}$ that is not in $\Phi$, we retrieve its second-order neighbor sequence from \emph{tgt-Dict} of the current batch.
If successful, we obtain the second-order neighbor information of $src_0$ using merely the sampled first-order neighbors with negligible storage costs (since $m$ is small).
Then, we update the corresponding interactive information of the node not in $\Phi$ with its higher-order neighbors, reconstructing a new sequence for the source node as $Q^{src_0}_{new}$.
The same operation is applied to each node in the target sequence.
Finally, the higher-order neighbor information obtained through successful retrieval provides more interactive information between source and target nodes for both bipartite and non-bipartite graphs.

For instance, under a bipartite graph setting, $Q^{src_0}$ and $Q^{tgt_0}$ are $\{tgt_0, tgt_1, tgt_2, tgt_0\}$ and $\{src_0, src_1, src_2, src_0\}$, respectively.
We focus on the nodes in $\Psi-\Phi$ but do not process nodes $src_0$ and $tgt_0$ themselves.
For non-bipartite graphs, the encoding method is the same, but with fewer nodes in $\Psi-\Phi$ that need to be processed.
Assuming that nodes $src_1$ and $tgt_2$ are in the current batch dictionaries with first-order neighbor sets $\{tgt_1,tgt_3\}$ and $\{src_2,src_4\}$, we can obtain $Q^{src_0}_{new}$ and $Q^{tgt_0}_{new}$ after reconstructing, as shown in \cref{equ:new sequence}.

\begin{equation}\label{equ:new sequence}
\begin{aligned}
    &Q^{src_0}_{new}=\{tgt_0,tgt_1,\bcancel{tgt_2}\rightarrow\{src_2,src_4\},tgt_0\},\\
    &Q^{tgt_0}_{new}=\{src_0,\bcancel{src_1}\rightarrow\{tgt_1,tgt_3\},src_2,src_0\}.
\end{aligned}
\end{equation}

For the node pair $src_0$ and $tgt_0$, the number of interactions in $Q^{src_0}$/$Q^{tgt_0}$ is $[2,2]$.
For nodes $src_1$ and $src_2$, the numbers of interactions in $Q^{src_0}_{new}$/$Q^{tgt_0}$ are $[0,1]$ and $[1,1]$.
For nodes $tgt_1$ and $tgt_2$, the numbers of interactions in $Q^{src_0}$/$Q^{tgt_0}_{new}$ are $[1,1]$ and $[1,0]$.
Therefore, the interactive information of node $src_0$ and $tgt_0$ are denoted as $I^{src_0}=[[2,2],[1,1],[1,0],[2,2]]^{\top}$ and $I^{tgt_0}=[[2,2],[0,1],[1,1],[2,2]]^{\top}$.
Following \cite{yu2023towards,tian2023freedyg}, we finally obtain the BIE embeddings $\mathcal{B}^{src_0}\in\mathbb{R}^{n\times d_{b}}$ and $\mathcal{B}^{tgt_0}\in\mathbb{R}^{n\times d_{b}}$ by 
\begin{equation}\label{equ:BIE}
\begin{aligned}
    &\mathcal{B}^{src_0}=Linear(ReLU(Linear(I^{src_0}))),\\
    &\mathcal{B}^{tgt_0}=Linear(ReLU(Linear(I^{tgt_0}))).
\end{aligned}
\end{equation}
where $d_b$ denotes the dimension of interaction embeddings.
This interaction encoding module significantly improves the utilization of interactive dynamics by increasing the receptive fields of nodes without additional high computational costs.

\header{\textbf{Seasonality \& Trend Encoding (STE).}}
The change of node interaction patterns over time in dynamic graphs contains rich information. 
For example, some nodes interact stably with fixed nodes, showing a steady interaction pattern over time, which is easy to model. 
Conversely, other nodes are easily affected by seasonal changes or social hot spots in the real world, showing rolling interaction patterns, which remain challenging to learn. 
Besides, most previous dynamic graph models consider the temporal features and the interactive information independently, failing to collaboratively model these dynamics and overlooking the importance of capturing seasonality and trend features that are common and informative in time-series data like dynamic graphs.

Inspired by the idea of decomposition in dealing with time-series data \cite{anderson1976time,cleveland6seasonal,wu2021autoformer}, we decompose the interaction patterns of nodes in dynamic graphs over time into the seasonal and trend parts, making it easier to model complex patterns from two views.
The seasonal part represents the periodic fluctuations, while the trend part refers to the long-term, consistent tendency.
Specifically, for an interaction sequence $Q$, we apply an average pooling operation to smooth out fluctuations in $Q$ to obtain the trend feature $\mathcal{Z}^{t}\in\mathbb{R}^{n\times d_{tr}}$, and then figure out the seasonal part $\mathcal{Z}^{s}\in\mathbb{R}^{n\times d_{s}}$ by filtering out the trend feature, where $d_{tr}$ and $d_{s}$ denote the dimensions of trend and seasonal embeddings, respectively.
After decomposition, $\mathcal{Z}^{t}$ indicates the smoothed overall trend, and $\mathcal{Z}^{s}$ reflects the potential seasonal variation of the temporal interaction pattern, undisturbed by the effect of the trend.
Mathematically, for a chronologically ordered interaction sequence $Q$ of a given node, that is, the normalized index of the interacted neighbor nodes, the decomposition process is shown in \cref{equ:seasonal and trend embedding}.
\begin{equation}\label{equ:seasonal and trend embedding}
\begin{aligned}
    &\mathcal{Z}^{t}=\text{AvgPool}(Q_{padded}),\\
    &\mathcal{Z}^{s}=Q-\mathcal{Z}^{t}.
\end{aligned}
\end{equation}
where $Q_{padded}$ denotes $Q$ after padding to maintain the sequence length unchanged. 
The illustration of the process is shown in \cref{fig:pipeline}.
This encoding module jointly models temporal and interactive information and effectively captures potential changes of interaction patterns over time through a simple decomposition.

\header{\textbf{Transformer Architecture at Interaction Level.}}
As introduced in \cref{three SAMs}, the input sequence $H_{IL}\in\mathbb{R}^{n\times(d+d_{i})}$ of our SAM at IL includes the initial sequence $\mathcal{H}\in\mathbb{R}^{n\times d}$ and the well-designed temporal-interactive dynamics embedding $\mathcal{I}\in\mathbb{R}^{n\times d_{i}}$. 
The embedding $\mathcal{I}$ consists of three parts: the MTE embedding $\mathcal{T}^{mix}\in\mathbb{R}^{n\times d_{t}}$, the BIE embedding $\mathcal{B}\in\mathbb{R}^{n\times d_{b}}$, and STE embedding $\mathcal{Z}=[\mathcal{Z}^{s},\mathcal{Z}^{t}]\in\mathbb{R}^{n\times (d_{s}+d_{tr})}$.
Based on these encoding modules, we concatenate all embeddings of the source node and the target node, respectively,
\begin{equation}\label{equ:final embedding}
\begin{aligned}
    &{H}_{IL}^{src}=\mathcal{H}^{src}||\mathcal{T}^{mix,src}|| \mathcal{B}^{src}||\mathcal{Z}^{src}\in\mathbb{R}^{n\times(d+d_{i})},\\
    &{H}_{IL}^{tgt}=\mathcal{H}^{tgt}||\mathcal{T}^{mix,tgt}|| \mathcal{B}^{tgt}||\mathcal{Z}^{tgt}\in\mathbb{R}^{n\times(d+d_{i})}.
\end{aligned}
\end{equation}
where $d=d_n+d_e$ and $d_{i}=d_{t}+d_b+d_s+d_{tr}$. 
Then, ${H}_{IL}^{src}$ and ${H}_{IL}^{tgt}$ are fed into Transformer, respectively.
In practice, we adopt multi-head SAM (MSA) with $J$ heads.
For $1\leq j\leq J$, the calculation of the attention in \cref{equ:vanilla attention} is updated as shown in \cref{equ:mha}.
\begin{equation}\label{equ:mha}
\begin{aligned}
    &Head_j=Softmax\left(\frac{({H}_{IL}W_{Q_j})({H}_{IL}W_{K_j})^{\top}}{\sqrt{d_{k}}}\right){H}_{IL}W_{V_j},\\
    &MSA({H}_{IL}) = (Head_1||\ldots||Head_j||\ldots||Head_{J})W_O,
\end{aligned}
\end{equation}
where $W_{Q_j},W_{K_j},W_{V_j}\in\mathbb{R}^{(d+d_{i})\times d_k}$ and $W_O\in\mathbb{R}^{(d+d_{i})\times (d+d_{i})}$ are learnable projection matrices.
We then use a feed-forward network (FFN) with residual connection and layer normalization (LN) to get the output of the $\ell$-th layer ${H}_{IL}^{\ell}\in\mathbb{R}^{n\times(d+d_{i})}$ as shown in \cref{equ:final}
\begin{equation}\label{equ:final}
\begin{aligned}
    &{H}_{IL}^{\prime\ell}=LN({H}_{IL}^{\ell-1}+MSA({H}_{IL}^{\ell-1})),\\
    &{H}_{IL}^{\ell}=LN({H}_{IL}^{\prime\ell}+FFN({H}_{IL}^{\prime\ell})).
\end{aligned}
\end{equation}

The final embeddings of the source node and the target node after $L$ layers are denoted as ${H}_{IL}^{L,src}\in\mathbb{R}^{n\times(d+d_{i})}$ and ${H}_{IL}^{L,tgt}\in\mathbb{R}^{n\times(d+d_{i})}$, which are used for downstream tasks.

\subsection{Time Complexity Analysis}\label{sec:short time complexity}
Let $n$ denote the length of the sampled neighbor sequence. 
The time complexity of TIDFormer is analyzed in three components as follows:
(1) For \textbf{MTE} and \textbf{STE}, the time complexities of the two modules are both $O(n)+O(n)=O(n)$, which includes fine \& coarse temporal embeddings, as well as the AvgPool and subtraction operations over a sequence with $n$ nodes, respectively.
This indicates that computing our proposed MTE and STE embeddings does not change the overall time complexity, since they are simple linear operations.
(2) For \textbf{BIE}, based on first-order sampling, we introduce a bidirectional mechanism to reconstruct the sequence.
This process consists of: (i) First-order sampling, $O(n)$, and (ii) Selective second-order retrieval and reconstruction, $O(n)$, with $n$ nodes in the worst case and constant time hash lookups.
Thus, the time complexity of BIE is $O(n)+O(n)=O(n)$. 
In contrast, prior approaches such as TGAT and TGN employ second-order sampling when stacking layers, leading to a significantly higher complexity of $O(n^2)$. 
(3) The \textbf{Transformer backbone} in TIDFormer has the same time complexity as vanilla Transformers $O(n^2 \times dim)$, where $dim$ denotes the dimension of the hidden features.

Overall, the time complexity of TIDFormer is $O(n)+O(n^2\times dim)$, which is comparable to previous first-order sampling Transformer-based methods ($O(n)+O(n^{2}\times dim)$), while demonstrating superior performance.
Furthermore, TIDFormer exhibits significant efficiency and performance advantages over earlier second-order sampling Transformer-based methods($O(n^{2})+O(n^{2}\times dim)$), as $n\ll n^{2}$.
These results are empirically validated in \cref{app:Efficiency Experiments}.

\section{Experiment}
In this section, we verify the effectiveness and efficiency of TIDFormer across a variety of dynamic graph datasets.
We provide a series of comprehensive and detailed analyses, demonstrating the superior experimental results of our proposed model.

\begin{table*}[t]
\centering
\caption{AP (\%) for dynamic link prediction on real-world datasets under transductive setting with random, historical, and inductive negative sampling strategies (NSS).}
\label{tab:ap-trans-full}
\resizebox{1.00\textwidth}{!}
{
\setlength{\tabcolsep}{0.6mm}
{
\begin{tabular}{l|c|ccccccccccc}
\toprule
NSS                    & Datasets    & JODIE        & DyRep        & TGAT         & TGN          & CAWN         & GraphMixer   & DyGFormer& FreeDyG& RepeatMixer& TPNet&TIDFormer\\ \midrule
\multirow{8}{*}{rnd}  & Wikipedia   & 96.50 $\pm$ 0.14 & 94.86 $\pm$ 0.06 & 96.94 $\pm$ 0.06 & 98.45 $\pm$ 0.06 & 98.76 $\pm$ 0.03 & 97.25 $\pm$ 0.03 & 99.03 $\pm$ 0.02  & 99.26 $\pm$ 0.01 & 99.16 $\pm$ 0.02&        \textbf{99.32 $\pm$ 0.03}& \underline{99.30 $\pm$ 0.03}\\
                       & Reddit      & 98.31 $\pm$ 0.14 & 98.22 $\pm$ 0.04 & 98.52 $\pm$ 0.02 & 98.63 $\pm$ 0.06 & 99.11 $\pm$ 0.01 & 97.31 $\pm$ 0.01 & 99.22 $\pm$ 0.01  & \textbf{99.48 $\pm$ 0.01}& 99.22 $\pm$ 0.01& 99.27 $\pm$ 0.00& \underline{99.38 $\pm$ 0.03}\\
                       & MOOC        & 80.23 $\pm$ 2.44 & 81.97 $\pm$ 0.49 & 85.84 $\pm$ 0.15 & 89.15 $\pm$ 1.60 & 80.15 $\pm$ 0.25 & 82.78 $\pm$ 0.15 & 87.52 $\pm$ 0.49  & 89.61 $\pm$ 0.19 & 92.76 $\pm$ 0.10& \textbf{96.39 $\pm$ 0.09}& \underline{93.42 $\pm$ 0.15}\\
                       & LastFM      & 70.85 $\pm$ 2.13 & 71.92 $\pm$ 2.21 & 73.42 $\pm$ 0.21 & 77.07 $\pm$ 3.97 & 86.99 $\pm$ 0.06 & 75.61 $\pm$ 0.24 & 93.00 $\pm$ 0.12  & 92.15 $\pm$ 0.16 & 94.14 $\pm$ 0.06& \underline{94.50 $\pm$ 0.08}& \textbf{94.60 $\pm$ 0.17}\\
                       & Enron       & 84.77 $\pm$ 0.30 & 82.38 $\pm$ 3.36 & 71.12 $\pm$ 0.97 & 86.53 $\pm$ 1.11 & 89.56 $\pm$ 0.09 & 82.25 $\pm$ 0.16 & 92.47 $\pm$ 0.12  & 92.51 $\pm$ 0.05 & 92.66 $\pm$ 0.07& \underline{92.90 $\pm$ 0.17}& \textbf{93.17 $\pm$ 0.06}\\
                       & Social Evo. & 89.89 $\pm$ 0.55 & 88.87 $\pm$ 0.30 & 93.16 $\pm$ 0.17 & 93.57 $\pm$ 0.17 & 84.96 $\pm$ 0.09 & 93.37 $\pm$ 0.07 & {94.73 $\pm$ 0.01}& \textbf{94.91 $\pm$ 0.01}& 94.72 ± 0.02& 94.73 $\pm$ 0.02& \underline{94.76 $\pm$ 0.06}\\
                       & UCI         & 89.43 $\pm$ 1.09 & 65.14 $\pm$ 2.30 & 79.63 $\pm$ 0.70 & 92.34 $\pm$ 1.04 & 95.18 $\pm$ 0.06 & 93.25 $\pm$ 0.57 & 95.79 $\pm$ 0.17 & 96.28 $\pm$ 0.11 & 96.74 $\pm$ 0.08& \underline{97.35 $\pm$ 0.04}& \textbf{97.44 $\pm$ 0.10}\\
                        \cline{2-13}
                        & Avg. Rank   & 9.43 & 10.00 & 8.86 & 6.71 & 7.43 & 8.43 & 4.57 &                         3.14 & 3.71 & 2.14 &\textbf{1.57} \\
                       \midrule
\multirow{8}{*}{hist} & Wikipedia   & 83.01 $\pm$ 0.66 & 79.93 $\pm$ 0.56 & 87.38 $\pm$ 0.22 & 86.86 $\pm$ 0.33 & 71.21 $\pm$ 1.67 & 90.90 $\pm$ 0.10 & 82.23 $\pm$ 2.54  & \textbf{91.59 $\pm$ 0.57}& 90.20 $\pm$ 1.04& 81.55 $\pm$ 4.10
& \underline{91.21 $\pm$ 0.78}\\
                       & Reddit      & 80.03 $\pm$ 0.36 & 79.83 $\pm$ 0.31 & 79.55 $\pm$ 0.20 & 81.22 $\pm$ 0.61 & 80.82 $\pm$ 0.45 & 78.44 $\pm$ 0.18 & {81.57 $\pm$ 0.67}  & \textbf{85.67 $\pm$ 1.01} & 83.02 $\pm$ 1.20& 81.02 $\pm$ 1.31& \underline{83.27 $\pm$ 1.01}\\
                       & MOOC        & 78.94 $\pm$ 1.25 & 75.60 $\pm$ 1.12 & 82.19 $\pm$ 0.62 & {87.06 $\pm$ 1.93} & 74.05 $\pm$ 0.95 & 77.77 $\pm$ 0.92 & 85.85 $\pm$ 0.66  & 86.71 $\pm$ 0.81 & 92.19 $\pm$ 0.58& \underline{92.69 $\pm$ 0.95}& \textbf{95.37 $\pm$ 1.09}\\
                       & LastFM      & 74.35 $\pm$ 3.81 & 74.92 $\pm$ 2.46 & 71.59 $\pm$ 0.24 & 76.87 $\pm$ 4.64 & 69.86 $\pm$ 0.43 & 72.47 $\pm$ 0.49 & {81.57 $\pm$ 0.48}  & 79.71 $\pm$ 0.51 & 86.73 $\pm$ 0.34& \underline{87.74 $\pm$ 0.50}& \textbf{88.68 $\pm$ 0.57}\\
                       & Enron       & 69.85 $\pm$ 2.70 & 71.19 $\pm$ 2.76 & 64.07 $\pm$ 1.05 & 73.91 $\pm$ 1.76 & 64.73 $\pm$ 0.36 & 77.98 $\pm$ 0.92 & 75.63 $\pm$ 0.73  & {78.87 $\pm$ 0.82} & \textbf{87.38 $\pm$ 0.18}& 80.79 $\pm$ 1.68& \underline{81.59 $\pm$ 1.17}\\
                       & Social Evo. & 87.44 $\pm$ 6.78 & 93.29 $\pm$ 0.43 & 95.01 $\pm$ 0.44 & 94.45 $\pm$ 0.56 & 85.53 $\pm$ 0.38 & 94.93 $\pm$ 0.31 & {97.38 $\pm$ 0.14}  & \textbf{97.79 $\pm$ 0.23} & \underline{97.40 ± 0.15}& 96.80 $\pm$ 0.41& {97.39 $\pm$ 0.67}\\
                       & UCI         & 75.24 $\pm$ 5.80 & 55.10 $\pm$ 3.14 & 68.27 $\pm$ 1.37 & 80.43 $\pm$ 2.12 & 65.30 $\pm$ 0.43 & 84.11 $\pm$ 1.35 & 82.17 $\pm$ 0.82  & {86.10 $\pm$ 1.19} & \underline{87.23 $\pm$ 0.23}& 86.34 $\pm$ 0.80& \textbf{89.57 $\pm$ 1.17}\\ 
                       \cline{2-13}
                        & Avg. Rank   & 8.14 & 9.00 & 8.14 & 6.00 & 10.00 & 6.86 & 6.43 & 3.00 & 2.57 & 4.14 & \textbf{1.71} \\
                       \midrule
\multirow{8}{*}{ind}  & Wikipedia   & 75.65 $\pm$ 0.79 & 70.21 $\pm$ 1.58 & 87.00 $\pm$ 0.16 & 85.62 $\pm$ 0.44 & 74.06 $\pm$ 2.62 & 88.59 $\pm$ 0.17 & 78.29 $\pm$ 5.38  & \underline{90.05 $\pm$ 0.79} & 88.86 $\pm$ 0.97& 79.35 $\pm$ 5.52
& \textbf{91.15 $\pm$ 1.17}\\
                       & Reddit      & 86.98 $\pm$ 0.16 & 86.30 $\pm$ 0.26 & 89.59 $\pm$ 0.24 & 88.10 $\pm$ 0.24 & \textbf{91.67 $\pm$ 0.24} & 85.26 $\pm$ 0.11 & 91.11 $\pm$ 0.40  & 90.74 $\pm$ 0.17 & 91.11 $\pm$ 0.73& 88.19 $\pm$ 0.33& \underline{91.57 $\pm$ 0.50}\\
                       & MOOC        & 65.23 $\pm$ 2.19 & 61.66 $\pm$ 0.95 & 75.95 $\pm$ 0.64 & 77.50 $\pm$ 2.91 & 73.51 $\pm$ 0.94 & 74.27 $\pm$ 0.92 & 81.24 $\pm$ 0.69  & {83.01 $\pm$ 0.87} & 83.11 $\pm$ 1.28& \textbf{88.18 $\pm$ 0.97}& \underline{84.53 $\pm$ 1.11}\\
                       & LastFM      & 62.67 $\pm$ 4.49 & 64.41 $\pm$ 2.70 & 71.13 $\pm$ 0.17 & 65.95 $\pm$ 5.98 & 67.48 $\pm$ 0.77 & 68.12 $\pm$ 0.33 & 73.97 $\pm$ 0.50  & 72.19 $\pm$ 0.24 & 75.46 $\pm$ 0.78& \textbf{77.99 $\pm$ 1.30}& \underline{75.69 $\pm$ 1.17}\\
                       & Enron       & 68.96 $\pm$ 0.98 & 67.79 $\pm$ 1.53 & 63.94 $\pm$ 1.36 & 70.89 $\pm$ 2.72 & 75.15 $\pm$ 0.58 & 75.01 $\pm$ 0.79 & 77.41 $\pm$ 0.89  & {77.81 $\pm$ 0.65} & \textbf{83.17 $\pm$ 0.50}& 75.36 $\pm$ 1.81& \underline{79.67 $\pm$ 0.66}\\
                       & Social Evo. & 89.82 $\pm$ 4.11 & 93.28 $\pm$ 0.48 & 94.84 $\pm$ 0.44 & 95.13 $\pm$ 0.56 & 88.32 $\pm$ 0.27 & 94.72 $\pm$ 0.33 & \underline{97.68 $\pm$ 0.10}  & {97.57 $\pm$ 0.15} & 97.52 ± 0.28& 97.35 $\pm$ 0.32& \textbf{97.70 $\pm$ 0.36}\\
                       & UCI         & 65.99 $\pm$ 1.40 & 54.79 $\pm$ 1.76 & 68.67 $\pm$ 0.84 & 70.94 $\pm$ 0.71 & 64.61 $\pm$ 0.48 & 80.10 $\pm$ 0.51 & 72.25 $\pm$ 1.71  & {82.35 $\pm$ 0.73} & \underline{84.20 $\pm$ 0.34}& 77.26 $\pm$ 1.57& \textbf{85.81 $\pm$ 1.39}\\
                       \cline{2-13}
                        & Avg. Rank   & 9.57 & 10.29 & 7.14 & 7.14 & 7.86 & 7.00 & 4.57 & 3.57 & 2.71 & 4.43 & \textbf{1.57} \\  
                    \bottomrule
\end{tabular}
}
}
\end{table*}

\begin{table}[t]
\centering
\caption{AUC-ROC (\%) for dynamic node classification.}
\label{tab:auc-node-class}
\resizebox{0.4\textwidth}{!}
{
\setlength{\tabcolsep}{1.5mm}
{
\begin{tabular}{l|cc|c}
\toprule
Methods    & Wikipedia                & Reddit                   & Avg. Rank \\ 
\midrule
JODIE      & \textbf{88.99 $\pm$ 1.05} & 60.37 $\pm$ 2.58          & 4.50 
\\
DyRep      & 86.39 $\pm$ 0.98          & 63.72 $\pm$ 1.32          & 5.50 
\\
TGAT       & 84.09 $\pm$ 1.27          & \textbf{70.04 $\pm$ 1.09} & 4.50 
\\
TGN        & 86.38 $\pm$ 2.34          & 63.27 $\pm$ 0.90          & 6.50 
\\
CAWN       & 84.88 $\pm$ 1.33          & 66.34 $\pm$ 1.78          & 5.50 
\\
GraphMixer & 86.80 $\pm$ 0.79          & 64.22 $\pm$ 3.32          & 4.50 
\\
DyGFormer  & 87.44 $\pm$ 1.08          & 68.00 $\pm$ 1.74          & 3.00 
\\ 
TIDFormer  & \underline{87.53 $\pm$ 1.12} & \underline{69.59 $\pm$ 1.70}       &                                              \textbf{2.00} \\ 
\bottomrule
\end{tabular}
}
}
\end{table}

\subsection{Experimental Setup}
\textbf{Datasets and Baselines.}
We conduct experiments on seven real-world datasets, including Wikipedia, Reddit, MOOC, LastFM, Enron, Social Evo., and UCI, spanning various domains.
The detailed statistics of these datasets are given in \cref{app:stat}.
Following previous works \cite{cong2023do,yu2023towards}, we use the same $70\%/15\%/15\%$ chronological splits to generate the train/validation/test sets for each dataset.
To better validate the effectiveness of our model, we compare TIDFormer to several strong baselines, including
JODIE \cite{DBLP:conf/kdd/KumarZL19}, DyRep \cite{DBLP:conf/iclr/TrivediFBZ19}, TGAT \cite{DBLP:conf/iclr/XuRKKA20}, TGN \cite{DBLP:journals/corr/abs-2006-10637}, CAWN \cite{DBLP:conf/iclr/WangCLL021}, GraphMixer \cite{cong2023do}, DyGFormer \cite{yu2023towards}, FreeDyG \cite{tian2023freedyg}, RepeatMixer \cite{zou2024repeat} and TPNet \cite{lu2024improving}.

\header{\textbf{Implementation Details and Evaluation Metrics.}}
Consistent with earlier works \cite{DBLP:conf/iclr/XuRKKA20,DBLP:journals/corr/abs-2006-10637}, we select two common tasks: dynamic link prediction and node classification.
For link prediction, we evaluate TIDFormer under both transductive and inductive settings using three negative sampling strategies \cite{poursafaei2022towards}.
The transductive setting focuses on predicting future links between nodes observed during training, while the inductive setting aims to predict links between unseen nodes.
The three negative sampling strategies include random (rnd), historical (hist), and inductive (ind) negative sampling strategies, which makes the link prediction task more challenging for evaluation and testing.
For more implementation details and hyperparameter configuration, please refer to \cref{app:config}.
Regarding evaluation metrics, we use Average Precision (AP) for link prediction. 
For node classification, we use Area Under the Receiver Operating Characteristic Curve (AUC-ROC) to address the label imbalance in dynamic graph datasets.

\subsection{Experimental Result}
\textbf{Performance Comparison to Baselines.}
The performances of TIDFormer and other previous SOTA models in dynamic link prediction on seven real-world datasets under transductive and inductive settings with three negative sampling strategies on AP are shown in \cref{tab:ap-trans-full} and \cref{tab:ap-ind}.
The best and the runner-up results are highlighted in bold and underlined fonts.

We observe that TIDFormer performs excellently, outperforming SOTA models on most datasets under both transductive and inductive settings with three negative sampling strategies.
For instance, TIDFormer achieves an average rank of $1.57$ for transductive dynamic link prediction with a random negative sampling strategy on AP, clearly demonstrating its superiority.
Additionally, compared to the latest SOTA models, such as DyGFormer and FreeDyG, the significant performance improvements of TIDformer on both bipartite and non-bipartite datasets illustrate that the proposed BIE module notably enhances the utilization of interactive dynamics without direct sampling of higher-order neighbors.
Meanwhile, the MTE module and the STE module improve the encoding process of both temporal and interactive features, leading to better model performances under most experimental settings.

Furthermore, the performances of TIDFormer and other previous models in dynamic node classification on AUC-ROC are shown in \cref{tab:auc-node-class}.
TIDFormer outperforms most previous baselines, with an average ranking of $2$, which further verifies the effectiveness in exploiting temporal and interactive dynamics in TIDFormer.

\begin{figure*}[t]
    \centering
    \includegraphics[width=1.9\columnwidth]{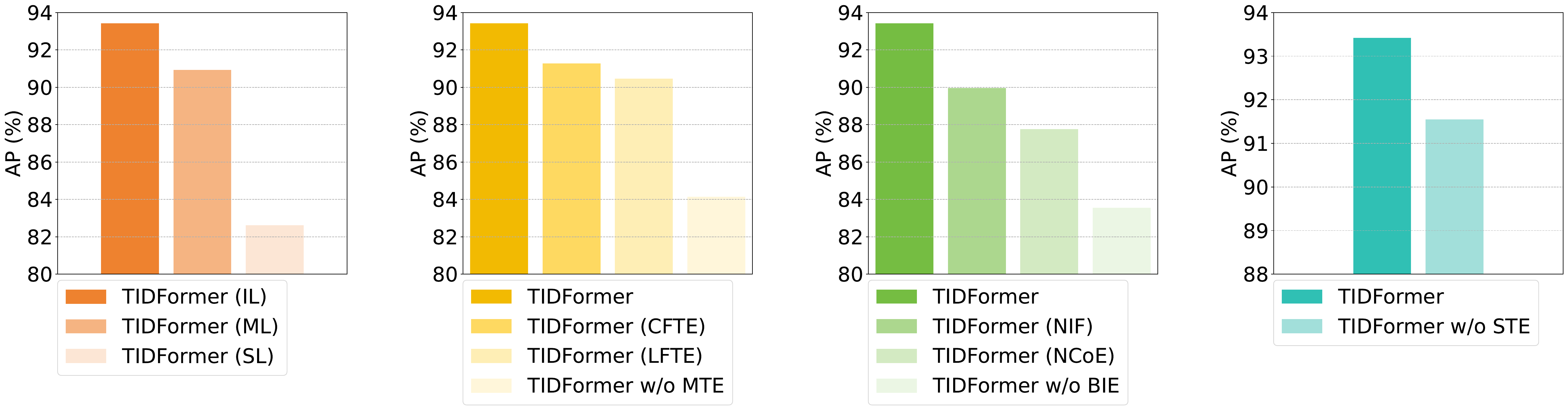}
    \caption{The results of the ablation experiments on MOOC. 
    Orange: Comparison of three types of SAMs.
    Yellow: Performance of TIDFormer w/ and w/o different temporal encoding modules.
    Green: Model performance w/ and w/o various neighbor and interaction encoding modules.
    Cyan: Performance of TIDFormer w/ and w/o seasonality \& trend encoding.
    }
    \label{fig:ab_mooc}
\end{figure*}

\subsection{Ablation Study}
To further validate the effectiveness of our proposed SAM at IL and the three modules for encoding temporal and interactive dynamics in TIDFormer, we conduct extensive ablation experiments on each component.
The results on MOOC are reported in \cref{fig:ab_mooc}.
Please refer to \cref{app:ab} for more results on other datasets.

For the SAM, we reconstruct TIDFormer with different SAMs at SL, ML, and IL and then evaluate their performance.
The superiority of TIDFormer with the SAM at IL over others verifies the inadequacy of the SAM at SL in capturing interactive information, as well as the negative impact on model performance caused by the mixed chronological sequence in the SAM at ML.
For the encoding modules, we compare the performance of the standard TIDFormer to versions of TIDFormer without MTE, BIE, or STE, respectively. 
For a fair and detailed comparison, we also provide the results after replacing each module in TIDFormer with traditional modules used in previous SOTA models. 
This includes replacing the MTE module with the constant or learnable fine-grained time encoding (CFTE or LFTE) module commonly used in baselines such as GraphMixer, DyGFormer, and FreeDyG, and replacing the BIE module with the neighbor co-occurrence encoding (NCoE) module from DyGFormer and the node interaction frequency (NIF) encoding module from FreeDyG.
The results show that the standard TIDFormer, fully equipped with all encoding modules, performs best across all experiments.
Either removing any module or replacing it with a traditional module used in previous SOTA models decreases the model performance to some extent. 
This highlights the necessity of the SAM at IL and the proposed encoding modules in TIDFormer.

\begin{figure}[b]
\vskip -0.1in
    \centering
    \includegraphics[width=0.99\columnwidth]{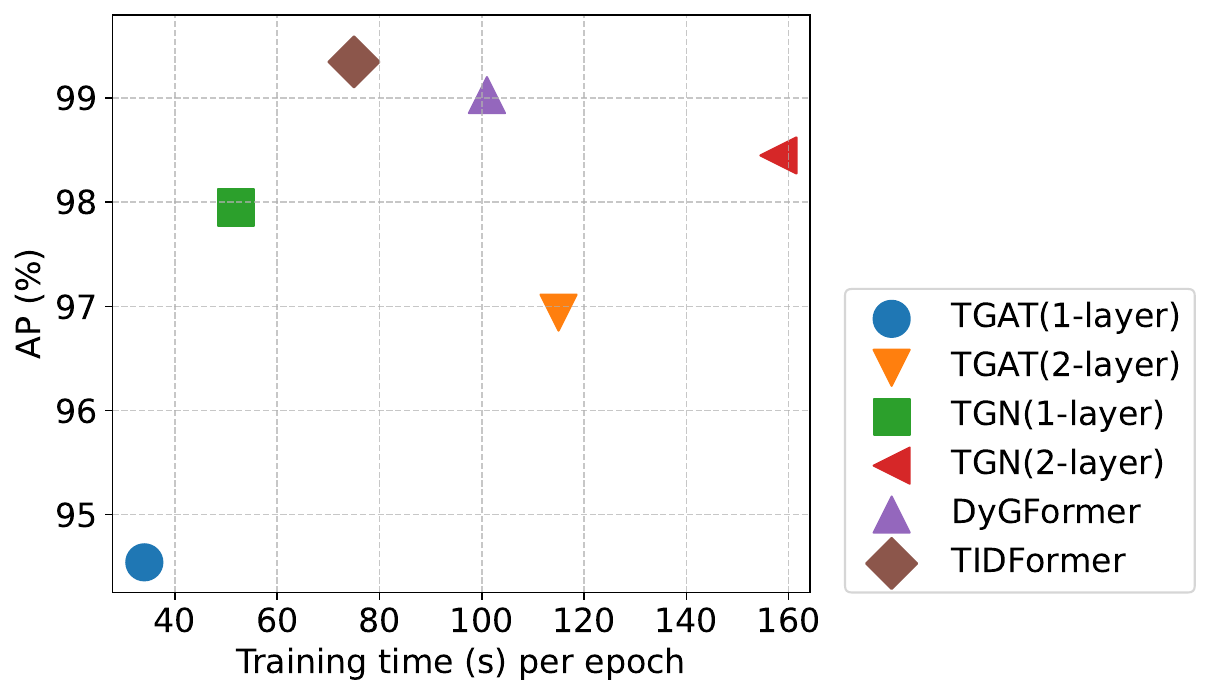}
    \vskip -0.1in
    \caption{The AP (\%) and training time (s) per epoch in Wikipedia dataset. The closer to the upper left corner, the better effectiveness and efficiency of the model.}
    \label{fig:training_time}
\end{figure}

\subsection{Efficiency Analysis}\label{app:Efficiency Experiments}
We conduct time efficiency experiments to analyze the training time per epoch of Transformer-based baselines, including TGAT (1-layer/2-layer) \cite{DBLP:conf/iclr/XuRKKA20}, TGN (1-layer/2-layer) \cite{DBLP:journals/corr/abs-2006-10637}, DyGFormer \cite{yu2023towards}, and TIDFormer, using the Wikipedia dataset. 
The results, presented in \cref{fig:training_time}, show that our proposed model achieves the best link prediction precision while demonstrating superior computational efficiency compared to all standard Transformer-based baselines.
Although TGAT and TGN exhibit greater efficiency with a one-layer architecture (i.e., first-order sampling), the corresponding link prediction performance suffers significant degradation. 
This highlights TIDFormer's success in balancing model performance and computational cost.
Nonetheless, we plan to further optimize the model architecture in terms of scalability and computational efficiency in future work, particularly in comparison to memory-based models. 
Potential solutions include exploring linear Transformers or developing more efficient sampling algorithms.

\section{Conclusion}
In this work, we propose TIDFormer, a Transformer-based dynamic graph model that fully mines temporal and interactive dynamics.
To address the open problem of uninterpretable definitions of the SAM on dynamic graphs, we first clarify its specific structure at the interaction level and experimentally verify its interpretability.
Our proposed TIDFormer is equipped with three simple yet effective encoding modules: the MTE module, the BIE module, and the STE module.
These modules comprehensively capture temporal and interactive dynamics using merely the sampled first-order neighbors.
We conduct extensive experiments on various dynamic graph datasets to verify the effectiveness and efficiency of TIDFormer. 
The experimental results indicate that TIDFormer excels, outperforming SOTA models across most datasets and experimental settings.

\begin{acks}
This research was supported in part by National Natural Science Foundation of China (No. 92470128, No. U2241212), by Beijing Outstanding Young Scientist Program No.BJJWZYJH012019100020098, by Huawei-Renmin University joint program on Information Retrieval. We also wish to acknowledge the support provided by the fund for building world-class universities (disciplines) of Renmin University of China, by Engineering Research Center of Next-Generation Intelligent Search and Recommendation, Ministry of Education, by Intelligent Social Governance Interdisciplinary Platform, Major Innovation \& Planning Interdisciplinary Platform for the “Double-First Class” Initiative, Public Policy and Decision-making Research Lab, and Public Computing Cloud, Renmin University of China. The work was partially done at Gaoling School of Artificial Intelligence, Beijing Key Laboratory of Research on Large Models and Intelligent Governance, MOE Key Laboratory of Data Engineering and Knowledge Engineering, Engineering Research Center of Next-Generation Intelligent Search and Recommendation, MOE, and Pazhou Laboratory (Huangpu), Guangzhou, Guangdong 510555, China.
\end{acks}

\newpage
\twocolumn
\bibliographystyle{ACM-Reference-Format}

\bibliography{reference}


\begin{table*}[!t]
\centering
\caption{Statistics of the seven real-world datasets.}
\vskip -0.1in

\label{tab:7data_stat}
\resizebox{0.78\textwidth}{!}
{
\setlength{\tabcolsep}{0.45mm}
{
\begin{tabular}{l|cccccccc}
\toprule
Datasets    & Domains     & Nodes & Edges   & Dim. of Node\&Edge Feat & Bipartite & Duration  & Unique Steps & Time Granularity    \\ 
\midrule
Wikipedia   & Social      & 9,227  & 157,474   & -- \& 172   & True      & 1 month    &  152,757      &  Unix timestamps   \\
Reddit      & Social      & 10,984 & 672,447   & -- \& 172  & True      & 1 month    &  669,065      &  Unix timestamps          \\
MOOC        & Interaction & 7,144  & 411,749   & -- \& 4   & True      & 1 month    &  345,600      & Unix timestamps         \\
LastFM      & Interaction & 1,980  & 1,293,103 & -- \& --    & True      & 4.3 years    &   1,283,614     & Unix timestamps           \\
Enron       & Social      & 184    & 125,235   & -- \& --   & False     & 3.6 years    &   22,632     &    Unix timestamps        \\
Social Evo. & Proximity   & 74     & 2,099,519 & -- \& 2    & False     & 8 months    &  565,932      &  Unix timestamps         \\
UCI         & Social      & 1,899  & 59,835    & -- \& --    & False     & 196 days    &  58,911      &  Unix timestamps         \\
\bottomrule
\end{tabular}
}
}
\vskip -0.1in
\end{table*}

\newpage
\appendix
\section{Datasets and Baselines}
\subsection{Statistics of Datasets}\label{app:stat}
The detailed statistics of seven real-world datasets are illustrated in \cref{tab:7data_stat}. 
All the real-world datasets are publicly available\footnote{\url{https://zenodo.org/record/7213796\#.Y1cO6y8r30o}}.
However, we find discrepancies in the duration statistics of the MOOC, LastFM, and Enron datasets compared to those reported in previous work \cite{poursafaei2022towards,yu2023towards,tian2023freedyg} (17 months, 1 month, 3 years) versus our results (1 month, 4.3 years, and 3.6 years), despite obtaining the datasets from the same official website. 
The duration of the dataset affects the choice of calendar-based time segment in our proposed MTE module.
Therefore, we use the corrected time span as the basis for the selection of time segment in our experiments.

\section{Implementation Details}
\subsection{Model and Hyperparameter Configuration}\label{app:config}
The backbone model of TIDFormer is a 2-layer Transformer with 2 attention heads.
The dimension of the MTE embedding $d_t$, BIE embedding $d_b$, and STE embeddings $d_s+d_{tr}$ are 100, 50, and 100, respectively.
Before feeding into the Transformer, ${H}_{IL}$ will be dimensionically reduced through linear layers.
In the MTE module, the calendar information and the corresponding time segment $R$ are easily chosen according to the duration of each dataset.
Based on the duration statistics of seven real-world datasets (Wikipedia, Reddit, MOOC, LastFM, Enron, Social Evo., and UCI) as shown in \cref{tab:7data_stat}, the selected calendar information \& $R$ of them are as follows: weekly \& 4, weekly \& 4, weekly \& 4, yearly \& 5, yearly \& 4, monthly \& 8, and weekly \& 28, respectively.

To ensure a fair comparison between the latest SOTA works \cite{yu2023towards,tian2023freedyg,zou2024repeat,lu2024improving}, we implement TIDFormer under the framework of DyGLib \cite{yu2023towards} with several default parameters, such as setting the number of epochs to $100$ with an early stopping strategy. 
We employ Adam as the optimizer and adopt binary cross-entropy as the loss function during training. 
We save the model that achieves the best results on the validation set for final testing. 
Moreover, we conduct a grid search to identify the optimal settings for several critical hyperparameters with the random negative sampling strategy. 
The search ranges and relevant hyperparameters are detailed in \cref{tab:grid search}. 
To improve training efficiency, we process all datasets using a batch operation with a batch size of $200$. 
To ensure the accuracy of the experimental results, we conduct five runs of experiments using random seeds and report the final average values. 
Experiments are conducted on an NVIDIA RTX A6000 48GB GPU.
For reproducibility, we provide the source code of our proposed method at \url{https://github.com/Lucas-PJ/TIDFormer}.

\begin{table}[h]
\centering
\caption{Searched ranges of hyperparameters and the related methods.}
\vskip -0.1in
\label{tab:grid search}
{
\resizebox{0.48\textwidth}{!}
{
\begin{tabular}{c|c}
\toprule
{Hyperparameters}  & {Searched Ranges}  \\ 
\midrule
\begin{tabular}[c]{@{}c@{}}Number of Sampled First-order Neighbors\end{tabular}      & [16, 24, 32, 40, 48, 56, 64] \\

\begin{tabular}[c]{@{}c@{}}Neighbor Sampling Strategies\end{tabular}     & [recent, uniform] \\

Dropout Rate   & \begin{tabular}[c]{@{}c@{}}[0.0, 0.1, 0.2, 0.3, 0.4, 0.5]\end{tabular}  \\

\begin{tabular}[c]{@{}c@{}}Weight Decay\end{tabular} & \begin{tabular}[c]{@{}c@{}}[0.0, 1e-6, 1e-4]\end{tabular}   \\ 

Learning Rate   & \begin{tabular}[c]{@{}c@{}}[1e-3, 1e-4, 1e-5, 1e-6]\end{tabular}  \\

Dimension of Hidden Features   & \begin{tabular}[c]{@{}c@{}}[8, 16, 32, 64, 128]\end{tabular}  \\

\bottomrule
\end{tabular}
}

}
\vskip -0.1in
\end{table}

\section{Supplementary Experimental Results}
\subsection{Results on Link Prediction}

\begin{table*}[!htbp]
\centering
\caption{AP (\%) for dynamic link prediction on real-world datasets under inductive setting with random, historical, and inductive negative sampling strategies (NSS).}
\vskip -0.1in
\label{tab:ap-ind}
\resizebox{1.01\textwidth}{!}
{
\setlength{\tabcolsep}{0.6mm}
{
\begin{tabular}{l|c|ccccccccccc}
\toprule
NSS                    & Datasets    & JODIE        & DyRep        & TGAT         & TGN          & CAWN         & GraphMixer   & DyGFormer     & FreeDyG & RepeatMixer&TPNet&TIDFormer\\ 
\midrule
\multirow{8}{*}{rnd}  & Wikipedia   & 94.82 $\pm$ 0.20 & 92.43 $\pm$ 0.37 & 96.22 $\pm$ 0.07 & 97.83 $\pm$ 0.04 & 98.24 $\pm$ 0.03 & 96.65 $\pm$ 0.02 & 98.59 $\pm$ 0.03  & \textbf{98.97 $\pm$ 0.01}  & 98.70 $\pm$ 0.05&98.91 $\pm$ 0.01& \underline{98.93 $\pm$ 0.02}\\
                       & Reddit      & 96.50 $\pm$ 0.13 & 96.09 $\pm$ 0.11 & 97.09 $\pm$ 0.04 & 97.50 $\pm$ 0.07 & 98.62 $\pm$ 0.01 & 95.26 $\pm$ 0.02 & 98.84 $\pm$ 0.02  & \underline{98.91 $\pm$ 0.01}  & 98.85 $\pm$ 0.01&98.86 $\pm$ 0.01& \textbf{98.98 $\pm$ 0.03}\\
                       & MOOC        & 79.63 $\pm$ 1.92 & 81.07 $\pm$ 0.44 & 85.50 $\pm$ 0.19 & {89.04 $\pm$ 1.17} & 81.42 $\pm$ 0.24 & 81.41 $\pm$ 0.21 & 86.96 $\pm$ 0.43  & 87.75 $\pm$ 0.62  & 93.05 $\pm$ 0.12&\textbf{95.07 $\pm$ 0.26}& \underline{93.29 $\pm$ 0.39}\\
                       & LastFM      & 81.61 $\pm$ 3.82 & 83.02 $\pm$ 1.48 & 78.63 $\pm$ 0.31 & 81.45 $\pm$ 4.29 & 89.42 $\pm$ 0.07 & 82.11 $\pm$ 0.42 & 94.23 $\pm$ 0.09  & {94.89 $\pm$ 0.01}  & 94.98 $\pm$ 0.12&\underline{95.36 $\pm$ 0.11}& \textbf{95.41 $\pm$ 0.17}\\
                       & Enron       & 80.72 $\pm$ 1.39 & 74.55 $\pm$ 3.95 & 67.05 $\pm$ 1.51 & 77.94 $\pm$ 1.02 & 86.35 $\pm$ 0.51 & 75.88 $\pm$ 0.48 & {89.76 $\pm$ 0.34}  & {89.69 $\pm$ 0.17}  & 87.97 $\pm$ 0.29&\textbf{90.34 $\pm$ 0.28}& \underline{89.92 $\pm$ 0.45}\\
                       & Social Evo. & 91.96 $\pm$ 0.48 & 90.04 $\pm$ 0.47 & 91.41 $\pm$ 0.16 & 90.77 $\pm$ 0.86 & 79.94 $\pm$ 0.18 & 91.86 $\pm$ 0.06 & 93.14 $\pm$ 0.04  & \textbf{94.76 $\pm$ 0.05}  & 93.00 ± 0.11&93.24 $\pm$ 0.07& \underline{93.28 $\pm$ 0.10}\\
                       & UCI         & 79.86 $\pm$ 1.48 & 57.48 $\pm$ 1.87 & 79.54 $\pm$ 0.48 & 88.12 $\pm$ 2.05 & 92.73 $\pm$ 0.06 & 91.19 $\pm$ 0.42 & 94.54 $\pm$ 0.12  & {94.85 $\pm$ 0.10}  & 95.04 $\pm$ 0.12&\underline{95.74 $\pm$ 0.05}& \textbf{{95.95 $\pm$ 0.12}} \\
                     \cline{2-13}& Avg. Rank   & 8.71 & 9.86 & 9.14 & 7.71 & 7.00 & 8.43 & 4.71 & 2.86 &3.86 &2.14 &\textbf{1.57} \\ 
                       \midrule
\multirow{8}{*}{hist} & Wikipedia   & 68.69 $\pm$ 0.39 & 62.18 $\pm$ 1.27 & 84.17 $\pm$ 0.22 & 81.76 $\pm$ 0.32 & 67.27 $\pm$ 1.63 & \textbf{87.60 $\pm$ 0.30} & 71.42 $\pm$ 4.43  & 82.78 $\pm$ 0.30  & 84.11 $\pm$ 1.88&71.28 $\pm$ 4.33
& \underline{84.99 $\pm$ 1.89}\\
                       & Reddit      & 62.34 $\pm$ 0.54 & 61.60 $\pm$ 0.72 & 63.47 $\pm$ 0.36 & 64.85 $\pm$ 0.85 & 63.67 $\pm$ 0.41 & 64.50 $\pm$ 0.26 & {65.37 $\pm$ 0.60}  & \underline{66.02 $\pm$ 0.41}& \textbf{66.14 $\pm$ 1.40}&62.15 $\pm$ 1.72& {65.62 $\pm$ 0.92}\\
                       & MOOC        & 63.22 $\pm$ 1.55 & 62.93 $\pm$ 1.24 & 76.73 $\pm$ 0.29 & 77.07 $\pm$ 3.41 & 74.68 $\pm$ 0.68 & 74.00 $\pm$ 0.97 & 80.82 $\pm$ 0.30  & {81.63 $\pm$ 0.33}  & \underline{83.19 $\pm$ 1.72}&81.85 $\pm$ 1.60& \textbf{84.61 $\pm$ 1.17}\\
                       & LastFM      & 70.39 $\pm$ 4.31 & 71.45 $\pm$ 1.76 & 76.27 $\pm$ 0.25 & 66.65 $\pm$ 6.11 & 71.33 $\pm$ 0.47 & 76.42 $\pm$ 0.22 & 76.35 $\pm$ 0.52  & {77.28 $\pm$ 0.21}  & 81.12 $\pm$ 0.30&\underline{82.27 $\pm$ 1.22}& \textbf{82.79 $\pm$ 0.83}\\
                       & Enron       & 65.86 $\pm$ 3.71 & 62.08 $\pm$ 2.27 & 61.40 $\pm$ 1.31 & 62.91 $\pm$ 1.16 & 60.70 $\pm$ 0.36 & 72.37 $\pm$ 1.37 & 67.07 $\pm$ 0.62  & {{73.01 $\pm$ 0.88}}  & \textbf{82.86 $\pm$ 0.47}&74.60 $\pm$ 1.35& \underline{76.99 $\pm$ 1.21}\\
                       & Social Evo. & 88.51 $\pm$ 0.87 & 88.72 $\pm$ 1.10 & 93.97 $\pm$ 0.54 & 90.66 $\pm$ 1.62 & 79.83 $\pm$ 0.38 & 94.01 $\pm$ 0.47 & \textbf{96.82 $\pm$ 0.16}& {96.69 $\pm$ 0.14}  & 96.75 ± 0.31&96.38 $\pm$ 0.18& \underline{96.77 $\pm$ 0.19}\\
                       & UCI         & 63.11 $\pm$ 2.27 & 52.47 $\pm$ 2.06 & 70.52 $\pm$ 0.93 & 70.78 $\pm$ 0.78 & 64.54 $\pm$ 0.47 & 81.66 $\pm$ 0.49 & 72.13 $\pm$ 1.87  & {82.35 $\pm$ 0.39}  & \underline{85.52 $\pm$ 0.16}&78.48 $\pm$ 1.18& \textbf{86.11 $\pm$ 1.81}\\
                       \cline{2-13}& Avg. Rank   & 9.43 & 9.86 & 7.00 & 7.14 & 9.14 & 5.00 & 4.86 & 3.57 &2.29 &6.00 &\textbf{1.71} \\ 
                       \midrule
\multirow{8}{*}{ind}  & Wikipedia   & 68.70 $\pm$ 0.39 & 62.19 $\pm$ 1.28 & 84.17 $\pm$ 0.22 & 81.77 $\pm$ 0.32 & 67.24 $\pm$ 1.63 & \textbf{87.60 $\pm$ 0.29} & 71.42 $\pm$ 4.43  & \underline{87.54 $\pm$ 0.26}  & 84.10 $\pm$ 1.88&71.29 $\pm$ 4.33& {84.99 $\pm$ 1.89}\\
                       & Reddit      & 62.32 $\pm$ 0.54 & 61.58 $\pm$ 0.72 & 63.40 $\pm$ 0.36 & 64.84 $\pm$ 0.84 & 63.65 $\pm$ 0.41 & 64.49 $\pm$ 0.25 & {65.35 $\pm$ 0.60}  & 64.98 $\pm$ 0.20  & \textbf{66.13 $\pm$ 1.40}&62.14 $\pm$ 1.72& \underline{65.62 $\pm$ 0.92}\\
                       & MOOC        & 63.22 $\pm$ 1.55 & 62.92 $\pm$ 1.24 & 76.72 $\pm$ 0.30 & 77.07 $\pm$ 3.40 & 74.69 $\pm$ 0.68 & 73.99 $\pm$ 0.97 & 80.82 $\pm$ 0.30  & {81.41 $\pm$ 0.31}  & \underline{83.19 $\pm$ 1.72}&81.85 $\pm$ 1.60& \textbf{84.61 $\pm$ 1.17}\\
                       & LastFM      & 70.39 $\pm$ 4.31 & 71.45 $\pm$ 1.75 & 76.28 $\pm$ 0.25 & 69.46 $\pm$ 4.65 & 71.33 $\pm$ 0.47 & 76.42 $\pm$ 0.22 & 76.35 $\pm$ 0.52  & {77.01 $\pm$ 0.43}  & 81.12 $\pm$ 0.30&\underline{82.27 $\pm$ 1.22}& \textbf{82.79 $\pm$ 0.83}\\
                       & Enron       & 65.86 $\pm$ 3.71 & 62.08 $\pm$ 2.27 & 61.40 $\pm$ 1.30 & 62.90 $\pm$ 1.16 & 60.72 $\pm$ 0.36 & 72.37 $\pm$ 1.38 & 67.07 $\pm$ 0.62  & {72.85 $\pm$ 0.81}  & \textbf{82.87 $\pm$ 0.47}&74.60 $\pm$ 1.35& \underline{76.98 $\pm$ 1.21}\\
                       & Social Evo. & 88.51 $\pm$ 0.87 & 88.72 $\pm$ 1.10 & 93.97 $\pm$ 0.54 & 90.65 $\pm$ 1.62 & 79.83 $\pm$ 0.39 & 94.01 $\pm$ 0.47 & \underline{96.82 $\pm$ 0.17}& \textbf{96.91 $\pm$ 0.12}& 96.75 ± 0.31&96.38 $\pm$ 0.18& {96.78 $\pm$ 0.19}\\
                       & UCI         & 63.16 $\pm$ 2.27 & 52.47 $\pm$ 2.09 & 70.49 $\pm$ 0.93 & 70.73 $\pm$ 0.79 & 64.54 $\pm$ 0.47 & 81.64 $\pm$ 0.49 & 72.13 $\pm$ 1.86  & {82.06 $\pm$ 0.58}  & \underline{85.53 $\pm$ 0.16}&78.50 $\pm$ 1.18& \textbf{86.11 $\pm$ 1.81}\\
                       \cline{2-13}& Avg. Rank   & 9.43 & 9.86 & 7.29 & 7.29 & 9.29 & 5.14 & 5.00 & 3.14 &2.57 &5.14 &\textbf{1.86} \\ 
                       \midrule
\end{tabular}
}
}
\end{table*}

The results of AP for dynamic link prediction under transductive and inductive settings with three negative sampling strategies are reported in \cref{tab:ap-trans-full} and \cref{tab:ap-ind}. 
We conduct experiments to evaluate RepeatMixer on the Social Evo. dataset using its public code under the default settings, as RepeatMixer \cite{zou2024repeat} does not report results on the Social Evo. dataset or provide its optimal hyperparameter configuration.

\begin{figure*}[t]
    \centering
    \includegraphics[width=1.8\columnwidth]{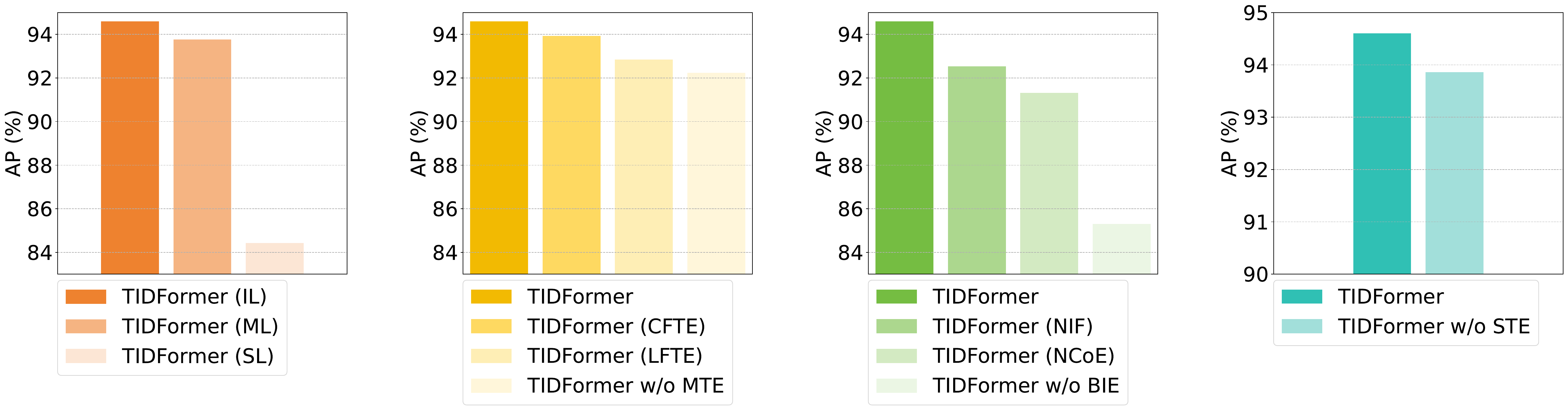}
    \vskip -0.1in
    \caption{The results of the ablation experiments on LastFM.}
    \label{fig:ab_last}
\end{figure*}

\begin{figure*}[t]
    \centering
    \includegraphics[width=1.8\columnwidth]{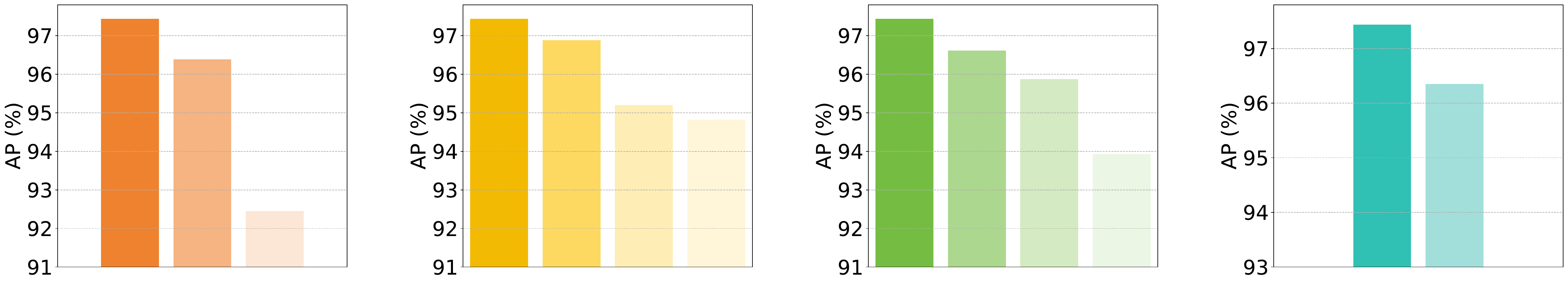}
    \vskip -0.1in
    \caption{The results of the ablation experiments on UCI. The color correspondences for the experimental settings are the same as in \cref{fig:ab_last}.}
    \label{fig:ab_uci}
\end{figure*}

\begin{figure*}[t]
    \centering
    \includegraphics[width=1.8\columnwidth]{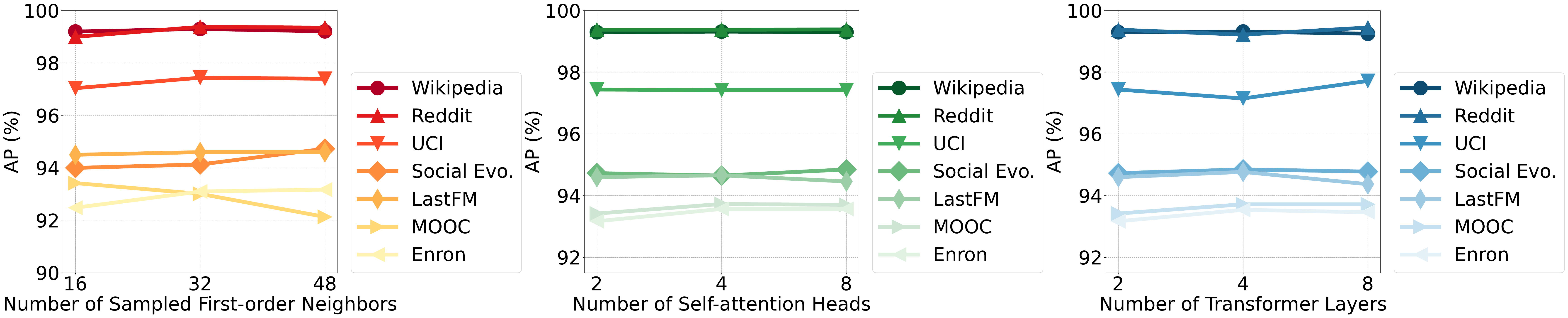}
    \vskip -0.1in
    \caption{The results of the hyperparameter study.}
    \label{fig:para}
    \vskip -0.5in
\end{figure*}

\subsection{Results on Ablation Experiments}\label{app:ab}
The results of the ablation experiments on LastFM and UCI are shown in \cref{fig:ab_last} and \cref{fig:ab_uci}, respectively.

\subsection{Results on Hyperparameter Study} \label{app:para}
We conduct detailed experiments on the selection of key hyperparameters in TIDFormer, including the number of sampled first-order neighbors, the number of Transformer layers, and the number of self-attention heads. 
These hyperparameters are crucial to the effectiveness and efficiency of TIDFormer. 
During the experiments, we adjust each hyperparameter while keeping the other parameters unchanged. 
Detailed experimental results are shown in \cref{fig:para}.

We observe that the optimal number of sampled first-order neighbors varies across different datasets. 
For instance, the Social Evo. dataset requires more neighbors mainly due to its complex and implicit temporal-interactive dynamics, whereas the MOOC dataset only needs to sample 16 first-order neighbors to achieve excellent performance. 
In addition, the number of Transformer layers and the number of self-attention heads have a slighter impact on model performance. 
In most cases, a 2-layer Transformer with 2 self-attention heads is sufficient to achieve ideal performance. 
However, increasing the number of model layers and self-attention heads will bring significant computational costs.
Therefore, it is necessary to trade off the computational efficiency and the model performance in practical experiments.

\end{document}